\documentclass[journal]{IEEEtran}
\usepackage{amsmath,amsfonts}
\usepackage{algorithm2e}
\usepackage{array}
\usepackage{textcomp}
\usepackage{stfloats}
\usepackage{url}
\usepackage{verbatim}
\usepackage{graphicx}
\usepackage{adjustbox}
\usepackage{cite}
\usepackage{hyperref}
\usepackage{url}
\usepackage{booktabs}
\usepackage{braket} 
\usepackage{soul}
\usepackage{placeins}
\usepackage[dvipsnames]{xcolor}
\usepackage{xurl} 
\usepackage{cleveref}

\usepackage{colortbl} 

\SetKwComment{Comment}{/* }{ */}
\RestyleAlgo{ruled}

\usepackage{xcolor} 

\usepackage[switch]{lineno}
\usepackage{multicol}
\usepackage{multirow}
\usepackage{mwe}
\usepackage{float}
\usepackage{balance}
\usepackage{lipsum}
\usepackage{balance}

\ifCLASSINFOpdf
\else
\fi

\begin{document}
\nolinenumbers

\title{Monitoring Post-Disaster Urban Recovery Using High-Resolution SAR Time Series and Unsupervised Learning: Evidence from the 2023 Türkiye–Syria Earthquake}

\author{
    \IEEEauthorblockN{Luigi Russo, \IEEEmembership{Student Member, IEEE}, 
    Deodato Tapete, \IEEEmembership{Member, IEEE}, 
    Silvia Liberata Ullo, \IEEEmembership{Senior Member, IEEE}, 
    and Paolo Gamba, \IEEEmembership{Fellow, IEEE}}
    
    \thanks{Luigi Russo and Paolo Gamba are with the Department of Electrical, Computer and Biomedical Engineering, University of Pavia, 27100 Pavia, Italy (e-mail: luigi.russo02@universitadipavia.it; paolo.gamba@unipv.it).}

    \thanks{Silvia Liberata Ullo is with the Department of Engineering, University of Sannio, 82100 Benevento, Italy (e-mail: ullo@unisannio.it).}

    \thanks{Deodato Tapete is with the Italian Space Agency (ASI), Via del Politecnico, 00133 Roma RM, Italia (e-mail: deodato.tapete@asi.it).}
}

\maketitle
 
\begin{abstract}
Monitoring the recovery phase after major natural disasters is essential for understanding how urban systems rebuild and progressively return to functionality. However, tracking reconstruction dynamics remains challenging due to the scarcity of reliable ground-truth information and the evolving nature of recovery processes. This paper proposes an unsupervised framework for post-disaster recovery monitoring based on multi-temporal SAR observations and deep learning (DL) anomaly detection. The method exploits COSMO-SkyMed (CSK) time series to identify persistent temporal anomalies associated with reconstruction activities and to derive spatially explicit recovery maps. The framework is applied to four cities severely affected by the 2023 Türkiye–Syria earthquakes, revealing heterogeneous reconstruction dynamics across different urban contexts. The results show spatially structured recovery patterns characterized by clusters of persistent anomalies associated with large-scale transformations such as reconstruction over previously damaged and cleared areas, the development of temporary container settlements, and the construction of new residential districts. A comparison with night-time lights (NTL) recovery indicators derived from SDGSAT-1 data highlights the complementary nature of the two observation modalities: while NTL signals reflect the restoration of electricity supply and nighttime socioeconomic activity, SAR-based anomaly signals capture structural transformations of the built environment and can reveal reconstruction processes at earlier stages. These results demonstrate that multi-temporal SAR observations combined with unsupervised learning provide an effective and operational approach for monitoring post-disaster reconstruction, particularly when labeled datasets describing recovery processes are unavailable. The proposed framework advances Earth Observation (EO) capabilities for long-term disaster recovery monitoring and provides a scalable methodology for tracking reconstruction dynamics in rapidly evolving post-disaster environments.
\end{abstract}


\IEEEpeerreviewmaketitle

\section{Introduction}
Large-scale natural disasters can produce severe and spatially heterogeneous impacts on the built-up environment, with cascading social, economic, and infrastructural consequences. In this context, Earth Observation (EO) has become a cornerstone for post-disaster analysis, enabling rapid, spatially explicit assessments over wide urban areas and providing an observational basis not only for immediate impact mapping, but also for tracking the longer trajectories of recovery and reconstruction.

A common feature across several hazard types is the concentration of physical forcing within a narrow temporal window, followed by prolonged phases of landscape alteration, recovery, and rebuilding. This class of temporally concentrated disasters includes sudden floods triggered by catastrophic failures (e.g., dam collapses) and large-scale landslides or debris flows. A paradigmatic example is the Vajont disaster in northern Italy (1963), where a massive rockslide into an artificial reservoir generated a catastrophic flood wave within minutes, causing extensive destruction and long-term socio-economic consequences \cite{Semenza2005,Kilburn2003}. Similar dynamics characterize rapid landslides and flash floods worldwide, where the destructive phase is confined to a short time window, while recovery unfolds over much longer temporal scales \cite{Haque2016,Petley2012}.

From an EO perspective, these dynamics translate into a key requirement: observations should capture the acute impact rapidly, yet remain consistent and repeatable over extended periods to monitor recovery-related transformations. Among remote sensing (RS) techniques, Synthetic Aperture Radar (SAR) is particularly valuable for disaster applications due to its all-weather, day--night imaging capability and its sensitivity to geometric and dielectric changes associated with damage. Over the last decade, advances in deep learning (DL) have further enhanced the exploitation of SAR for post-event mapping, improving both accuracy and timeliness when optical data are unavailable or severely limited \cite{Ge2020RSEReviewSARDamage}. Beyond single-scene change indicators, several DL-based approaches leverage the spatiotemporal information content of SAR acquisitions, most notably interferometric coherence time series, to infer damage patterns at scale shortly after an event \cite{Stephenson2021TGRS}. Methodological progress has been increasingly paired with strategies aimed at operational deployment, such as rapid domain adaptation across events and geographic contexts to mitigate domain shifts when labeled data are scarce \cite{Hertel2025SciRemoteSensingDomainAdapt}.

Despite major advances in rapid damage mapping, most EO studies remain concentrated on the early post-disaster phase and provide an essentially static snapshot of impacts shortly after the event. Post-disaster recovery is instead dynamic and heterogeneous, involving debris removal, temporary settlement installation, infrastructure repair, and progressive reconstruction unfolding over months or even years. Additionally, recovery and reconstruction are not always linear processes and frequently occur as series of spatially fragmented and time-wise disperse events. This implies that the effects can be seen over a very long time period and according to separate instances not happening at regular intervals. Systematic monitoring of this recovery phase is therefore critical for evaluating reconstruction progress and effectiveness, and informing long-term risk reduction strategies aligned with international frameworks such as the ``Build Back Better'' principle of the Sendai Framework for Disaster Risk Reduction \cite{UNDRR2015Sendai}. However, compared to damage assessment, EO-based recovery monitoring remains relatively limited and fragmented, partly because recovery signals are incremental, spatially heterogeneous, and temporally non-linear, and can be confounded by seasonal effects and background urban dynamics.

Early EO-based recovery studies largely relied on proxies of human activity, most notably satellite nighttime lights (NTL). Variations in radiance from the Defense Meteorological Satellite Program Operational Line-Scan System (DMSP--OLS) and, more recently, VIIRS NTL products, have been used to infer recovery dynamics after major disasters, including earthquake case studies such as Haiti (2010) and Wenchuan (2008) \cite{Li2019HaitiNTL,Wang2018Wenchuan}. With daily VIIRS time series, NTL-based analyses have progressed toward pixel-level characterization of recovery trajectories and revealed pronounced spatial heterogeneity and persistent inequalities within affected regions \cite{zheng2025nighttime}. Similar datasets have enabled spatially disaggregated assessments of power outages and restoration processes, as demonstrated for Puerto Rico after Hurricane Maria \cite{roman2019satellite,jia2023estimating}. Nonetheless, NTL remains an indirect indicator: radiance changes may reflect electricity availability, population return, emergency response, or reconstruction-related activities, while the effective spatial resolution of NTL products can obscure neighborhood-scale rebuilding patterns \cite{li2025npp}.

Beyond socio-economic proxies, recovery monitoring has increasingly explored indicators more directly linked to physical and environmental change. Passive microwave observations of vegetation optical depth (VOD) have been used to quantify post-fire regrowth across biomes, complementing optical vegetation indices \cite{bousquet2022monitoring}. More broadly, recent work has shifted toward high-resolution, multi-sensor EO time series, where optical and SAR acquisitions can capture land-cover transitions, debris clearance, reconstruction footprints, and the gradual reactivation of urban functions. Within the Committee on Earth Observation Satellites (CEOS) disaster community, long-term monitoring studies in Haiti after Hurricane Matthew have demonstrated that integrating optical and SAR time series can capture debris removal, reconstruction processes, and land-cover changes over multi-year periods \cite{de2021monitoring,velasquez2025monitoring}. In this respect, recent data-driven change analysis approaches, including self-supervised and reconstruction-based methods, provide promising tools to better separate recovery-related dynamics from background variability when labeled data are limited \cite{muzeau2022self,yadav2024unsupervised}.

While most of previous papers using EO observations to assess the recovery focused on situations where reconstruction was a relatively slow process, the increasing accessibility to high revisit time post-disaster time series can enable monitoring over urban and natural environments wherein efforts for reconstruction happen since the very first weeks. In this respect, the 2023 Turkiye-Syria earthquake sequence provides a particularly relevant benchmark to study both the opportunities and the limitations of EO-based post-disaster analysis at scale. On February 6, 2023, a Mw~7.8 earthquake struck southeastern Türkiye and northwestern Syria and was followed hours later by a Mw~7.5 event and thousands of aftershocks, substantially amplifying impacts on lives and infrastructure. Overall, the sequence caused more than 50{,}000 fatalities and widespread building damage across the region \cite{Yu2024NPJ}. Satellite-based damage products were generated within days, reflecting the maturity of operational crisis-mapping services and providing authoritative geospatial layers that are increasingly exploited as reference information for the development and validation of automated damage mapping \cite{CEMS2023Bulletin165,CEMS_EMSR648,CharterActivation797,UNOSAT2023EQReport}.

Within this context, several recent studies have exploited the 2023 earthquake as a large-scale testbed for data-driven damage assessment. Sun et al.~\cite{Sun2024} introduced the \href{https://github.com/ya0-sun/PostEQ-SARopt-BuildingDamage}{QuickQuakeBuildings} dataset, comprising over 4{,}000 damage-annotated buildings co-registered with post-event very-high-resolution (VHR) SAR and optical imagery, and highlighted the persistent performance gap between SAR-only and optical-only inputs. Complementary SAR-based approaches have explicitly exploited multitemporal coherence information. Using recurrent neural networks, Yang et al.~\cite{Yang2024EarthquakeSpectraRNN} modeled coherence time series to predict co-seismic coherence and derive damage proxy maps, achieving improved accuracy and reduced false-alarm rates compared to single-pair coherence analyses. In a subsequent study, the same authors extended this temporal perspective by introducing a Vision Transformer (ViT) trained on coherence time matrices that explicitly incorporate preseismic coherence, further improving damage discrimination and robustness, with demonstrated transferability across different seismic events \cite{Yang2024JAGViTCoherence}. In parallel, optical-based approaches have leveraged VHR imagery to map damage and collapse patterns. Xia et al.~\cite{Xia2023IJDRS} proposed a DL workflow for rapid damage assessment from ultra-high-resolution optical data, while Hac{\i}efendio{\u g}lu et al.~\cite{Haciefendioglu2024Buildings} addressed collapsed-building detection from post-disaster VHR optical imagery using semantic segmentation models.

Building on this line of work, our recent study \cite{Russo2026BDA} addressed post-event building damage assessment (BDA) using VHR SAR imagery from the COSMO-SkyMed (CSK) constellation, provided by the Italian Space Agency (ASI). The proposed framework integrates CSK observations with complementary exposure and vulnerability layers, including building stock information from the Global Earthquake Model (GEM), OpenStreetMap (OSM) building footprints with post-event annotations, and a high-resolution Digital Surface Model (DSM) from CartoSat-1 Euro-Maps. By embedding structural exposure, building typology, and 3D contextual information directly into the learning process, the approach improves damage discrimination and spatial consistency across multiple affected cities, and provides a methodological basis for extending EO-based analyses beyond the immediate post-event phase.

The widespread activation of satellite mapping services and the resulting multi-temporal EO coverage for the 2023 Türkiye--Syria earthquake have stimulated a growing body of recovery-oriented analyses based on satellite NTL. Early evidence from NASA's VIIRS Day/Night Band (DNB) revealed rapid post-event darkening patterns consistent with the mainshock footprint and enabled prompt delineation of impacted settlements, with independent confirmation using SDGSAT-1 nighttime observations \cite{levin2023using}. Subsequent studies quantified spatially explicit radiance changes from NOAA-20 VIIRS data and linked them to impact proxies, including population and building density as well as observed casualties, through spatial association analysis \cite{yuan2023changes}. Moving beyond immediate disruption mapping, recent work has explicitly targeted the recovery phase by introducing pixel-level loss metrics and composite NTL indicators to jointly characterize recovery performance and intrinsic resilience at subregional scales \cite{yang2025post}. Related analyses have further used NTL time series to describe the transition from post-event losses to reconstruction patterns, highlighting both strong impact gradients around the epicentral region and heterogeneous recovery rates across administrative levels \cite{xiao2024remote}.

A particularly relevant contribution is the study by Gong et al.~\cite{gong2025urban}, who exploited SDGSAT-1 NTL time series to cluster pixel-level recovery trajectories and reveal heterogeneous urban recovery patterns approximately one year after the event, providing a valuable benchmark for comparative recovery monitoring. At the same time, NTL-based products remain indirect proxies: radiance variations can reflect electricity availability and human activity, but they are only weakly and ambiguously linked to the progressive physical transformations of the built environment (e.g., debris clearance, temporary settlements, and reconstruction), especially at neighborhood scale. This motivates complementary recovery-monitoring approaches based on observables that are directly sensitive to physical change while remaining scalable when labeled reference data are limited.

In this paper, we pursue this direction by proposing a SAR-based framework for post-disaster recovery monitoring grounded in unsupervised analysis of SAR time series. Building on our previous damage-mapping work in the affected region \cite{Russo2026BDA}, we extend the focus from static impact assessment to the spatiotemporal characterization of recovery-related processes. Specifically, SAR temporal frames collected according to a regular acquisition plan are analysed using an unsupervised reconstruction-based model, where deviations from learned background dynamics are quantified through reconstruction errors. Persistent high-error responses are interpreted as anomaly signals, highlighting locations where significant surface changes occur and enabling the identification of recovery-related transformations over time. The resulting framework is designed to be operationally scalable and physically interpretable, allowing the systematic detection of heterogeneous recovery trajectories at urban-relevant spatial scales and facilitating direct comparison with NTL-based recovery benchmarks, including \cite{gong2025urban}.

The contributions of this work can be summarized as follows:
\begin{itemize}
    \item we propose an unsupervised framework for post-disaster recovery monitoring based on multi-temporal high resolution SAR time series and reconstruction-based anomaly detection. The method exploits reconstruction errors from a DL model to identify persistent temporal anomalies associated with recovery-related surface transformations, without requiring labeled training data.

    \item We derive spatially explicit recovery maps and \newline persistence-based recovery classes from cumulative SAR anomaly signals, enabling the characterization of heterogeneous reconstruction dynamics across different urban environments.

    \item We demonstrate the framework on four cities affected by the 2023 Türkiye--Syria earthquakes and analyse the spatial organization of recovery processes at the urban scale, including reconstruction over previously damaged areas, temporary container settlements, and newly developed residential districts.

    \item We compare SAR-derived recovery signals with independent NTL indicators derived from SDGSAT-1 data, highlighting the complementary information provided by structural change detection and nighttime socioeconomic activity proxies.
\end{itemize}

By addressing the largely unmet need for post-disaster recovery monitoring, this study aims to bridge the temporal gap in the disaster management cycle. The remainder of the paper is organized as follows. Section~\ref{study_area_and_data} describes the study area and the multi-temporal SAR datasets used in the analysis. Section~\ref{sec:methodology} presents the proposed unsupervised framework for recovery monitoring based on reconstruction-driven anomaly detection from SAR time series. Section~\ref{sec:results} reports the experimental results across the four cities affected by the 2023 Türkiye--Syria earthquake, analyzes the spatial organization of reconstruction dynamics, and compares the SAR-derived recovery signals with independent NTL indicators derived from SDGSAT-1 data. Section~\ref{sec:discussions} discusses the main implications of the results, highlights the complementary value of SAR and NTL observations for recovery monitoring, and outlines the main limitations of the proposed approach. Finally, Section~\ref{sec:conclusions} summarizes the main findings and identifies directions for future research.

\begin{figure*}
    \centering
    \includegraphics[width=\textwidth]{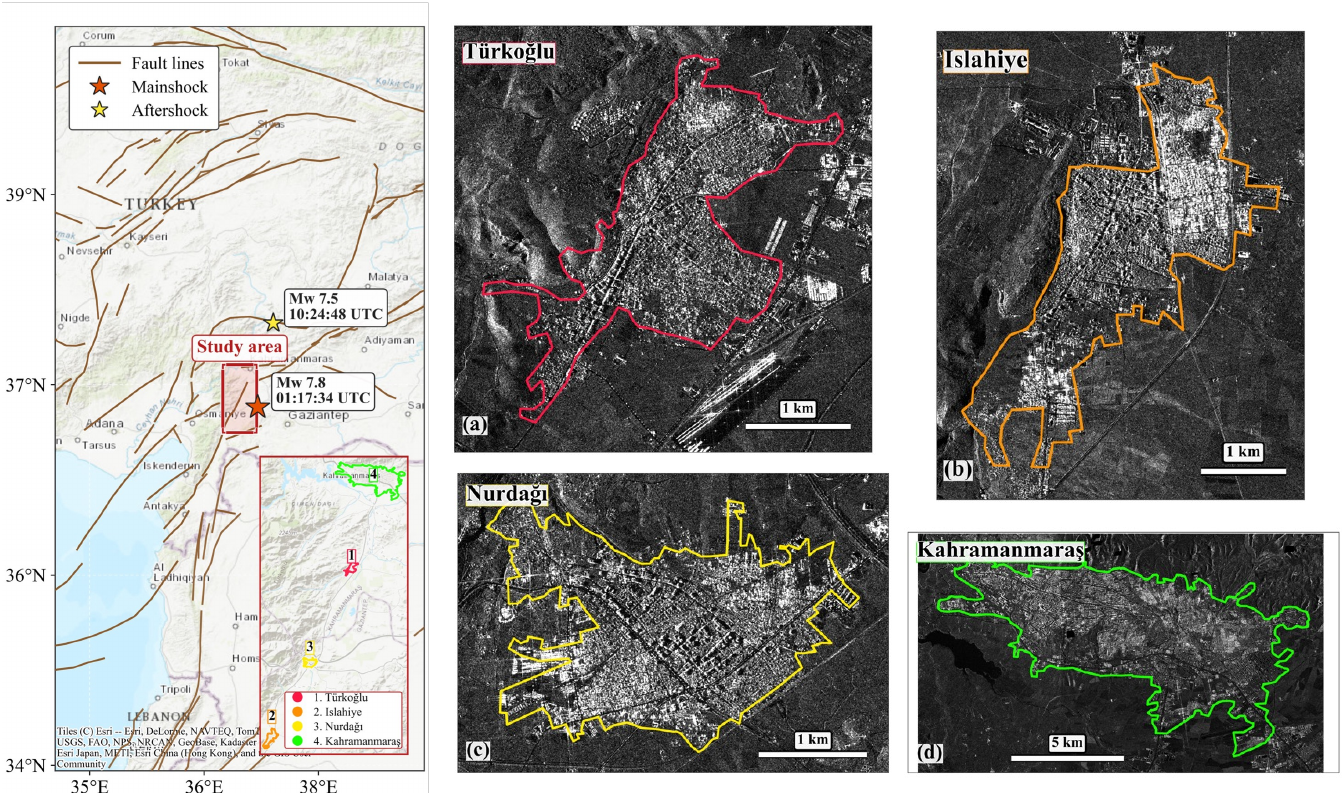}
    \caption{
    Study area and urban test sites affected by the 2023 Türkiye--Syria earthquake sequence.
    The left panel shows a regional overview of southeastern Türkiye, including major active fault systems (brown lines), the epicenters of the Mw~7.8 mainshock and Mw~7.5 aftershock (stars), and the spatial extent of the study area.
    Panels (a--d) show VHR CSK SAR imagery acquired immediately after the event for the four analyzed urban areas: Türkoğlu (a), Islahiye (b), Nurdağı (c), and Kahramanmaraş (d).
    Urban boundaries are overlaid to delineate the built-up areas considered in the analysis.
    Map lines delineate study areas and do not necessarily depict accepted national boundaries.
    }
    \label{fig:study_area}
\end{figure*}

\section{Study Area and Data}\label{study_area_and_data}

\subsection{Study Area}

This study focuses on urban areas affected by the 6 February 2023 Türkiye–Syria earthquake sequence, which caused widespread structural damage across southeastern Türkiye. A BDA based on VHR CSK SAR imagery was previously developed for this event, providing building-level damage severity and spatial distribution immediately after the earthquake. That work relied exclusively on post-event SAR observations and integrated complementary exposure and vulnerability information, including building footprints, DSM data, and GEM attributes, to improve damage discrimination, robustness, and cross-city generalization~\cite{Russo2026BDA}.

In the present study, those damage products serve as the baseline representation of the initial post-event conditions. Building on this foundation, the temporal scope of the observations is extended from single-date post-event acquisitions to multi-temporal CSK SAR time series, enabling the transition from static damage mapping to the monitoring of post-disaster recovery and reconstruction processes over medium- and longer-term horizons.

To ensure methodological consistency and allow a direct comparison between damage patterns and subsequent recovery dynamics, the analysis focuses on the same urban areas investigated in the previous BDA study. Four cities are considered: Nurdağı, Islahiye, Türkoğlu, and Kahramanmaraş. These cities lie close to the rupture zone of the Mw~7.8 and Mw~7.5 earthquake sequence and were among the most severely affected settlements in the epicentral region~\cite{USGS2023Event_us6000jllz}. They also span a geologically complex sector of southeastern Türkiye located along and near the interaction of major active fault systems, including the East Anatolian Fault, motivating subsequent analyses that relate spatial damage and recovery patterns to local ground conditions~\cite{Provost2024SciRepOffsets}. Their geographic context and spatial distribution are illustrated in Fig.~\ref{fig:study_area}.

\subsection{Data Sources}\label{sec:data_sources}

The proposed recovery-monitoring framework relies primarily on multi-temporal CSK SAR imagery, which provides VHR observations suitable for detecting structural changes in dense urban environments. SAR acquisitions are particularly valuable in post-disaster contexts because they are independent of illumination and weather conditions and are sensitive to variations in surface roughness, geometry, and dielectric properties. In this study, CSK time series constitute the main input to the DL-based methodology, enabling the derivation of an unsupervised change-detection product that tracks the temporal evolution of recovery-related transformations.

The CSK archive used in this analysis is provided by ASI and combines routine background acquisitions with targeted tasking performed after the earthquake within international disaster-mapping cooperation frameworks. In particular, additional acquisitions were made available through activities coordinated by the CEOS in support of the GEO Geohazard Supersites and Natural Laboratory (GSNL) initiative and the Kahramanmaraş Event Supersite EO data access mechanism~\cite{GSNL_Kahramanmaras_EOAccess}. This coordinated effort ensured the availability of temporally consistent SAR observations suitable for multi-temporal recovery monitoring over the years following the main seismic sequence.

Complementary datasets are used to contextualize damage and recovery patterns, support the interpretation of detected changes, and validate the consistency of the results. In particular, building damage maps from~\cite{Russo2026BDA} are employed as the reference baseline describing the immediate post-event damage state. 


In particular, these damage products provide the starting point for the recovery analysis, enabling comparison between the initial damage distribution and the spatial locations where recovery-related changes are subsequently detected.

High-resolution optical imagery available through Google Earth is used as the primary visual validation source for the unsupervised change patterns extracted from the CSK time series. Whenever temporally suitable scenes are available, Google Earth imagery enables direct inspection of debris removal, temporary facilities, construction sites, and newly built structures, providing an independent confirmation of the spatial plausibility of SAR-derived change signals. Sentinel-2 optical time series are additionally used as an auxiliary validation source in periods or locations where Google Earth imagery is temporally sparse or unavailable. Although characterized by lower spatial resolution, Sentinel-2 data provide very frequent coverage (every 5 days on average), enabling qualitative monitoring of medium-scale urban transformations and helping corroborate the timing and persistence of SAR-detected changes.

Geological maps at 1:100{,}000 scale provided by the Turkish General Directorate of Mineral Research and Exploration (MTA) are used to support the interpretation of recovery patterns in relation to local ground conditions. Since geological setting and site effects can influence the suitability of areas for reconstruction, this information allows us to investigate whether recovery activities and permanent rebuilding tend to concentrate on more stable lithological units, as opposed to geologically susceptible zones where seismic damage was previously widespread. In this way, the present experiment mocks the type of assessment that a technical officer from either a local or central administration may undertake once the EO-based recovery maps are available, and the purpose is to evaluate how much geologically resilient the reconstruction process could be (regardless if it is intentional and made consciously). 

Table~\ref{tab:csk_data} summarizes the CSK acquisitions used for the four study cities and highlights differences in acquisition geometry. Although a larger archive of SAR images is available for the post-earthquake period, this study focuses on a subset of representative acquisitions selected to capture major transitions in the recovery trajectory rather than to provide a uniformly dense temporal sampling, which could also be achieved, if needed, given the regular collection of CSK imagery. The selection is guided by visual evidence from Google Earth and Sentinel-2 time series in order to sample distinct phases of post-disaster recovery.

In particular, early acquisitions (February--May 2023) capture rapid post-event changes such as debris removal and the establishment of temporary settlements, while later acquisitions (from August 2023 onward) sample medium- and long-term reconstruction dynamics, including the emergence and consolidation of permanent residential buildings. This phase-oriented sampling strategy provides a practical compromise between temporal coverage and interpretability and supports the extraction of meaningful recovery-related change signals.

The three cities of Türkoğlu, Islahiye, and Nurdağı share the same Stripmap configuration (beam H4-22), resulting in identical sensing dates and perfectly aligned temporal sampling. In contrast, full coverage of Kahramanmaraş required two partially overlapping CSK SAR4 strips acquired with different imaging beams, including H4-03. This is due to the spatial extent and nearly east--west orientation of the urban footprint (Fig.~\ref{fig:study_area}d), which could not be fully covered within a single strip. As a result, part of the city is sampled on the same sensing dates as the other three cities, whereas the remaining portion is covered with only a slight temporal offset of around 4 days, as highlighted in Table~\ref{tab:csk_data}.

Despite this temporal offset, the acquisitions remain suitable for recovery monitoring. The consistent satellite platform and orbit direction preserve the radiometric and geometric homogeneity required for reconstruction-based anomaly detection, while the small date shifts do not affect the ability to capture the main recovery phases. Instead, the two acquisition sets can be interpreted as parallel temporal samplings of the same reconstruction timeline.

\begin{table*}[t]
\centering
\caption{COSMO-SkyMed SAR acquisitions used for post-earthquake recovery monitoring over the four study cities.}
\label{tab:csk_data}

\begin{adjustbox}{width=\textwidth}
\begin{tabular}{lllllll}
\hline
City & Satellite & Acquisition Mode & Beam & Polarization & Orbit & Sensing Dates \\
\hline

\multirow{6}{*}{Türkoğlu / Islahiye / Nurdağı}
& \multirow{6}{*}{CSK SAR4}
& \multirow{6}{*}{Stripmap HIMAGE}
& \multirow{6}{*}{H4-22}
& \multirow{6}{*}{HH}
& \multirow{6}{*}{Ascending}
& 25-Feb-2023 \\
&  &  &  &  &  & 16-May-2023 \\
&  &  &  &  &  & 20-Aug-2023 \\
&  &  &  &  &  & 11-Jan-2024 \\
&  &  &  &  &  & 03-Jun-2024 \\
&  &  &  &  &  & 25-Oct-2024 \\

\hline

\multirow{6}{*}{Kahramanmaraş}
& \multirow{6}{*}{CSK SAR4}
& \multirow{6}{*}{Stripmap HIMAGE}
& \multirow{6}{*}{H4-03}
& \multirow{6}{*}{HH}
& \multirow{6}{*}{Ascending}
& 01-Mar-2023 \\
&  &  &  &  &  & 20-May-2023 \\
&  &  &  &  &  & 24-Aug-2023 \\
&  &  &  &  &  & 15-Jan-2024 \\
&  &  &  &  &  & 07-Jun-2024 \\
&  &  &  &  &  & 29-Oct-2024 \\

\hline
\end{tabular}
\end{adjustbox}
\end{table*}

\section{Methodology}
\label{sec:methodology}

\subsection{Problem Formulation}
\label{subsec:problem_formulation}

Let $\{\mathbf{X}_t\}_{t=1}^{T}$ denote a multi-temporal SAR image sequence acquired over a given urban area, where each frame 
$\mathbf{X}_t \in \mathbb{R}^{H \times W}$ represents single-polarization (HH) SAR backscatter at acquisition time $t$. 
From the original images, we extract a set of spatio-temporal patches: 
\begin{equation}
\mathbf{x}^{(n)} \in \mathbb{R}^{T \times C \times H \times W}, \quad n=1,\dots,N ,
\end{equation}
where $C=1$ for single-polarization SAR, $H$ and $W$ denote the spatial height and width of each patch, respectively, and $N$ is the number of training samples.

The goal is to learn an \emph{unsupervised representation} of the expected spatio-temporal evolution of SAR backscatter and to identify deviations from this learned behavior. The choice of an unsupervised strategy is motivated by two complementary considerations.

First, the analysis targets the recovery phase of large-scale disasters, where up-to-date and spatially consistent reference information describing reconstruction activities is typically scarce or unavailable. As a result, the creation of reliable labels suitable for supervised learning is extremely challenging, particularly at city scale and over extended temporal windows.

Second, beyond data availability constraints, the unsupervised formulation reflects the intended operational scenario. In real-world post-disaster monitoring, EO time series data are often the primary and sometimes the only consistently available source of information. Therefore, the objective is to infer recovery-related changes directly from the SAR temporal evolution itself, without relying on external labels.

In this context, anomalies are defined as departures from the learned temporal patterns. In a post-disaster setting, temporally persistent anomalies are interpreted as signals consistent with recovery-related processes, including debris removal, temporary settlements, demolition, and permanent reconstruction.

\begin{figure*}
    \centering
    \includegraphics[width=\textwidth]{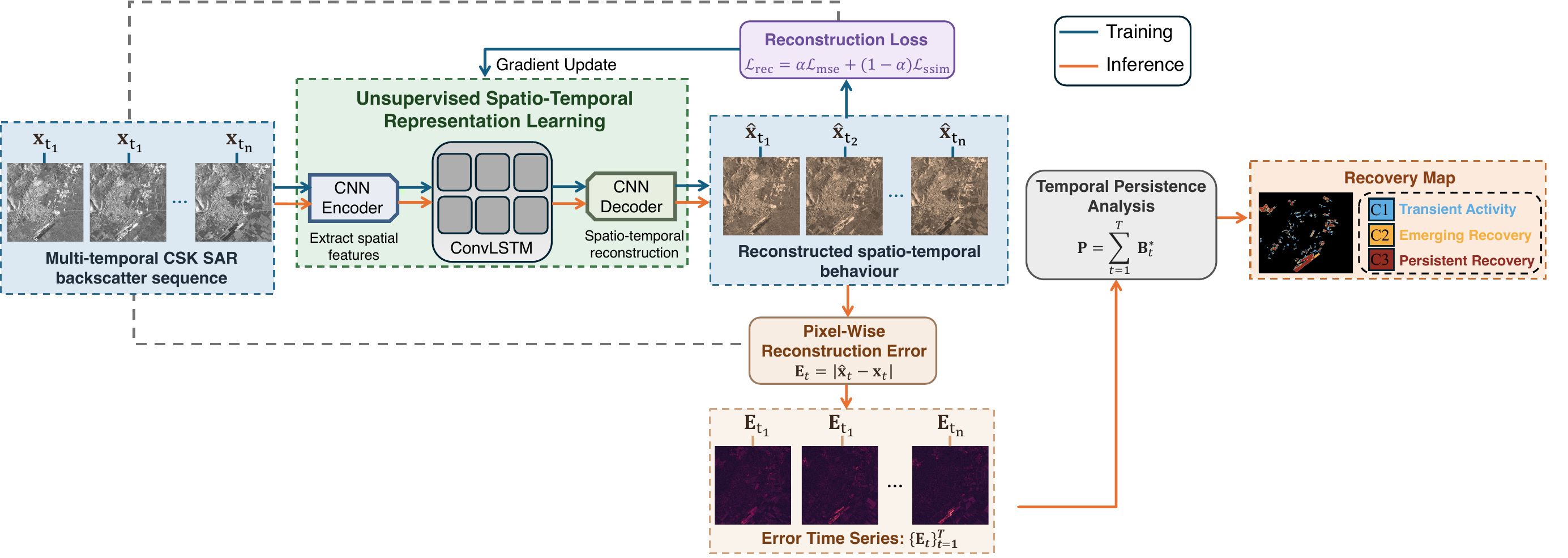}
    \caption{
    Unsupervised SAR-based framework for post-disaster recovery monitoring. A multi-temporal CSK SAR backscatter sequence is processed by a ConvLSTM-based autoencoder composed of a spatial CNN encoder, stacked ConvLSTM layers for latent temporal modeling, and a CNN decoder for sequence reconstruction. During training (blue arrows), the network is optimized using a reconstruction objective that balances radiometric fidelity and structural consistency. During inference (orange arrows), discrepancies between observed and reconstructed backscatter define anomaly signals and the associated reconstruction-error time series. After spatial refinement and temporal aggregation, recurrent anomalies are accumulated into a persistence surface and discretized into recovery categories (transient activity, emerging recovery, and persistent recovery), producing an operational recovery map of temporally consistent SAR-derived changes.
    }
    \label{fig:convlstm_workflow}
\end{figure*}

To capture the joint spatial and temporal structure of the SAR sequence, we adopt a ConvLSTM-based autoencoder (AE) architecture \cite{Shi2015ConvLSTM}. An overview of the overall workflow is provided in Fig.~\ref{fig:convlstm_workflow}. Formally, the model learns a mapping
\begin{equation}
\hat{\mathbf{x}}^{(n)} = 
\mathcal{D}\big( \mathcal{F}(\mathcal{E}(\mathbf{x}^{(n)})) \big),
\end{equation}
where $\mathcal{E}(\cdot)$ denotes a spatial encoder, $\mathcal{F}(\cdot)$ a stack of ConvLSTM layers modeling temporal dependencies in the latent space, and $\mathcal{D}(\cdot)$ a spatial decoder that reconstructs the input sequence.

The encoder $\mathcal{E}$ consists of strided convolutional layers that extract compact spatial representations from each frame. 
The encoded features are then processed by ConvLSTM layers, which extend classical LSTM units by replacing fully connected operations with convolutions, thereby preserving spatial structure while modeling temporal dynamics. 
Finally, the decoder $\mathcal{D}$ uses transposed convolutions to reconstruct the full spatio-temporal sequence.

The reconstruction error between $\mathbf{x}^{(n)}$ and $\hat{\mathbf{x}}^{(n)}$ is subsequently used as an anomaly score, forming the basis for the recovery-monitoring analysis presented in the following sections.

\paragraph{Spatial encoding}
Each frame $\mathbf{x}_t \in \mathbb{R}^{C \times h \times w}$ is mapped into a latent representation
$\mathbf{z}_t \in \mathbb{R}^{C_z \times h' \times w'}$:
\begin{equation}
\mathbf{z}_t = \mathcal{E}(\mathbf{x}_t),
\qquad t=1,\dots,T,
\label{eq:encoder}
\end{equation}
where $\mathcal{E}$ is the composition of convolutional layers with stride $2$ (thus $h' \approx h/4$ and
$w' \approx w/4$ in the present implementation) and a pointwise nonlinearity (ReLU).

\paragraph{ConvLSTM temporal modeling}
Given the latent sequence $\{\mathbf{z}_t\}_{t=1}^{T}$, temporal dependencies are modeled via ConvLSTM cells.
For a ConvLSTM layer, the gating mechanism is defined as:
\begin{align}
\mathbf{i}_t &= \sigma\big(\mathbf{W}_i * [\mathbf{z}_t, \mathbf{h}_{t-1}] + \mathbf{b}_i\big),\\
\mathbf{f}_t &= \sigma\big(\mathbf{W}_f * [\mathbf{z}_t, \mathbf{h}_{t-1}] + \mathbf{b}_f\big),\\
\mathbf{o}_t &= \sigma\big(\mathbf{W}_o * [\mathbf{z}_t, \mathbf{h}_{t-1}] + \mathbf{b}_o\big),\\
\mathbf{g}_t &= \tanh\big(\mathbf{W}_g * [\mathbf{z}_t, \mathbf{h}_{t-1}] + \mathbf{b}_g\big),
\label{eq:convlstm_gates}
\end{align}
where $*$ denotes convolution, $[\cdot,\cdot]$ concatenation along the channel dimension, and $\sigma(\cdot)$ the
logistic sigmoid. The memory and hidden states are updated as:
\begin{align}
\mathbf{c}_t &= \mathbf{f}_t \odot \mathbf{c}_{t-1} + \mathbf{i}_t \odot \mathbf{g}_t,\\
\mathbf{h}_t &= \mathbf{o}_t \odot \tanh(\mathbf{c}_t),
\label{eq:convlstm_state}
\end{align}
with $\odot$ denoting element-wise multiplication.

In our architecture, two ConvLSTM layers are stacked. The first layer receives $\mathbf{z}_t$ and outputs
$\mathbf{h}^{(1)}_t$, while the second layer receives $\mathbf{h}^{(1)}_t$ and outputs $\mathbf{h}^{(2)}_t$.
The final latent state $\mathbf{h}^{(2)}_T$ summarizes the sequence dynamics and is used for reconstruction.

\paragraph{Spatial decoding and reconstruction}
The decoder maps the final latent representation back to the image space:
\begin{equation}
\hat{\mathbf{x}}_t = \mathcal{D}(\mathbf{h}^{(2)}_T), \qquad t=1,\dots,T,
\label{eq:decoder}
\end{equation}
where $\mathcal{D}$ is implemented via transposed convolutions. In the current implementation, the same decoded
output is replicated across $t$ (i.e., no explicit autoregressive temporal decoder). This design still yields a
useful anomaly signal because the model is forced to encode sequence-consistent spatial patterns into
$\mathbf{h}^{(2)}_T$, and deviations from such patterns yield elevated reconstruction errors.

\subsection{Training Objective: Reconstruction Loss with MSE and SSIM}
\label{subsec:loss}

Given an input spatio-temporal patch $\mathbf{x}^{(n)}=\{\mathbf{x}^{(n)}_t\}_{t=1}^{T}$ and its reconstruction
$\hat{\mathbf{x}}^{(n)}=\{\hat{\mathbf{x}}^{(n)}_t\}_{t=1}^{T}$ produced by the ConvLSTM-based AE, the network is
trained by minimizing a reconstruction loss that combines a pixel-wise fidelity term (MSE) and a perceptual
structure term (SSIM). In practice, this encourages the model to reproduce both the radiometric content of SAR
patches and their spatial organization.

\paragraph{Mean squared error (MSE)}
The MSE penalizes point-wise deviations between the reconstructed and observed backscatter values. For a given
time step $t$, we define:
\begin{equation}
\mathcal{L}_{\mathrm{mse}}(t)=\frac{1}{hw}\sum_{i=1}^{h}\sum_{j=1}^{w}
\left(\hat{\mathbf{x}}_{t}(i,j)-\mathbf{x}_{t}(i,j)\right)^2,
\end{equation}
and the sequence-level MSE is obtained by averaging over time:
\begin{equation}
\mathcal{L}_{\mathrm{mse}}=\frac{1}{T}\sum_{t=1}^{T}\mathcal{L}_{\mathrm{mse}}(t).
\label{eq:mse}
\end{equation}
In our setting, minimizing $\mathcal{L}_{\mathrm{mse}}$ promotes accurate reconstruction of backscatter amplitudes,
thereby forcing the model to learn the dominant (``expected'') SAR dynamics and discouraging trivial solutions.

\paragraph{Structural similarity (SSIM)}
While MSE is sensitive to pixel-wise differences, it does not explicitly enforce preservation of local spatial
structure (e.g., edges, texture, and contrast), which are crucial in VHR SAR for capturing changes in urban
morphology. We therefore include SSIM, computed between $\mathbf{x}_t$ and $\hat{\mathbf{x}}_t$ on local windows
(or patch-wise, depending on implementation). For two corresponding windows $\mathbf{u}$ and $\mathbf{v}$, SSIM is:
\begin{equation}
\mathrm{SSIM}(\mathbf{u},\mathbf{v})=
\frac{(2\mu_u\mu_v + C_1)(2\sigma_{uv}+C_2)}
{(\mu_u^2+\mu_v^2+C_1)(\sigma_u^2+\sigma_v^2+C_2)},
\label{eq:ssim}
\end{equation}
where $\mu_u,\mu_v$ are local means, $\sigma_u^2,\sigma_v^2$ local variances, and $\sigma_{uv}$ the local
cross-covariance; $C_1$ and $C_2$ are small constants that stabilize the division. The SSIM loss is then:
\begin{equation}
\mathcal{L}_{\mathrm{ssim}}=\frac{1}{T}\sum_{t=1}^{T}\left(1-\mathrm{SSIM}(\hat{\mathbf{x}}_t,\mathbf{x}_t)\right).
\label{eq:ssim_loss}
\end{equation}
In our case, $\mathcal{L}_{\mathrm{ssim}}$ promotes reconstructions that preserve spatial organization of urban
backscatter patterns, reducing over-smoothing and improving the faithfulness of reconstructed built-up
structures. 

\paragraph{Combined objective.}
The final training loss is a convex combination of the two terms:
\begin{equation}
\mathcal{L}_{\mathrm{rec}}=\alpha\,\mathcal{L}_{\mathrm{mse}}+(1-\alpha)\,\mathcal{L}_{\mathrm{ssim}},
\label{eq:total_loss_nomask}
\end{equation}
with $\alpha\in[0,1]$ (in our implementation $\alpha=0.84$). This combination is motivated by the complementary
behavior of the two losses: MSE enforces radiometric fidelity at pixel level, whereas SSIM emphasizes local
structural consistency. Jointly optimizing them yields reconstructions that are both numerically accurate and
structurally plausible, which is desirable because the recovery-monitoring signal is ultimately derived from
reconstruction errors.


\subsection{Cross-city Training Protocol (Leave-One-City-Out)}
\label{subsec:loco}

To promote geographic generalization and operational robustness, we adopt a leave-one-city-out (LOCO) training protocol. In this setting, the model is trained using patches extracted from all cities except one, which is held out for validation. The procedure is repeated by iteratively selecting each city as the validation target.

This strategy is consistent with the experimental design adopted in our previous building damage assessment study on the 2023 Türkiye earthquake \cite{Russo2026BDA}, ensuring methodological continuity between the damage-mapping and recovery-monitoring phases. More importantly, it reflects a realistic operational scenario: recovery monitoring systems must be deployable in newly affected urban areas without retraining on city-specific data.

Urban morphology, building typologies, and SAR backscatter characteristics vary substantially across cities. A model trained and validated within the same urban context risks implicitly learning geographic or morphological biases rather than generalizable recovery-related dynamics. By withholding one city at a time, the LOCO protocol enforces cross-city transfer and evaluates whether the model can detect reconstruction-related anomalies independently of local geographic characteristics.

This design mirrors the rationale commonly adopted in cross-event damage assessment: just as damage models should generalize across different seismic contexts, recovery models should be capable of identifying reconstruction dynamics without being tailored to a specific urban layout. The LOCO strategy therefore serves both as a robustness test and as a proxy for real-world deployment conditions.

\subsection{Anomaly Scoring and Recovery-change Mapping}
\label{subsec:anomaly_mapping}

After training, the model is applied to the target city to obtain reconstructed sequences 
$\{\hat{\mathbf{X}}_t\}_{t=1}^{T}$. A per-pixel anomaly score is computed as the absolute reconstruction error:

\begin{equation}
\mathbf{E}_t(\mathbf{p}) = 
\left| \hat{\mathbf{X}}_t(\mathbf{p}) - \mathbf{X}_t(\mathbf{p}) \right|,
\qquad \mathbf{p}\in\Omega,
\label{eq:error_map}
\end{equation}

where $\Omega$ denotes the spatial domain.

Inference is performed in a patch-based manner over overlapping tiles. 
To reduce border artifacts during mosaicking, each patch contribution is weighted using a Hann window~\cite{Pielawski_2020} before aggregation. 
The final reconstruction error time series is obtained by weighted averaging of overlapping regions, producing a cube 
$\mathbf{E} \in \mathbb{R}^{T \times H \times W}$. 

\paragraph{Temporal persistence and spatial regularization}

Let $\mathbf{E}_t$ denote the reconstruction error map at time $t$.
For each frame $t$, the error map is first spatially smoothed:

\begin{equation}
\tilde{\mathbf{E}}_t = \mathcal{G}_\sigma(\mathbf{E}_t),
\end{equation}

where $\mathcal{G}_\sigma(\cdot)$ denotes Gaussian filtering with standard deviation $\sigma$.

An anomaly mask is then obtained through percentile-based thresholding:

\begin{equation}
\mathbf{B}_t(\mathbf{p}) =
\mathbb{I}\left(
\tilde{\mathbf{E}}_t(\mathbf{p}) >
\mathrm{perc}_{90}\big(\tilde{\mathbf{E}}_t\big)
\right),
\qquad \mathbf{p} \in \Omega,
\end{equation}

where $\mathrm{perc}_{90}(\cdot)$ denotes the 90th percentile of the smoothed error distribution and 
$\mathbb{I}(\cdot)$ is the indicator function defined as:

\begin{equation}
\mathbb{I}(A) =
\begin{cases}
1 & \text{if } A \text{ is true}, \\
0 & \text{otherwise}.
\end{cases}
\end{equation}

Therefore, $\mathbf{B}_t(\mathbf{p})$ is a binary map taking value 1 for pixels whose anomaly score exceeds the 90th percentile, and 0 otherwise.

Each $\mathbf{B}_t$ undergoes morphological refinement to enforce spatial coherence:

\begin{equation}
\mathbf{B}_t^{*} =
\mathcal{R}\big(
\mathcal{C}_{r_2}(
\mathcal{O}_{r_1}(\mathbf{B}_t)
)
\big),
\end{equation}

where $\mathcal{O}_{r_1}(\cdot)$ and $\mathcal{C}_{r_2}(\cdot)$ denote morphological opening and closing
using disk-shaped structuring elements of radius $r_1$ and $r_2$, respectively, and 
$\mathcal{R}(\cdot)$ removes connected components smaller than a predefined minimum area.

Temporal persistence is then computed as:

\begin{equation}
\mathbf{P}(\mathbf{p}) =
\sum_{t=1}^{T}
\mathbf{B}_t^{*}(\mathbf{p}),
\end{equation}

where $T$ denotes the total number of temporal frames.
The resulting persistence surface $\mathbf{P}$ measures how many times a pixel is detected as anomalous across the observation period.

Higher values of $\mathbf{P}(\mathbf{p})$ indicate temporally consistent transformations,
whereas lower values correspond to sporadic or short-lived anomaly activations.

The persistence surface $\mathbf{P}$ thus provides a quantitative measure of the temporal
recurrence of anomalous behavior at pixel level. However, to derive an operational
and interpretable recovery product, this continuous measure must be translated
into a discrete set of recovery categories.

\subsection{Recovery Class Definition and Post-processing}
\label{subsec:recovery_classes}

Building upon the persistence surface $\mathbf{P}$ introduced in the previous subsection, 
we convert the continuous anomaly recurrence measure into a discrete and operational 
recovery map. The underlying rationale is that temporally consistent deviations from the 
learned SAR dynamics are more likely to reflect structured recovery-related processes 
than isolated anomaly activations.

To this end, we define three recovery levels based on the temporal recurrence of refined anomaly masks $\mathbf{B}_t^{*}$. The classification rules and their operational definition are summarized in Table~\ref{tab:recovery_classes}.

The persistence value $P(\mathbf{p})$ therefore acts as a proxy for the temporal 
stability of anomalous behavior at pixel level. Pixels detected as anomalous only 
once are classified as \emph{Transient activity} (C1), representing isolated or 
short-lived deviations. Pixels exhibiting intermediate recurrence 
($2 \leq P < \tau$) are labeled as \emph{Emerging recovery} (C2), indicating 
areas undergoing repeated but not yet dominant transformations. Finally, pixels 
whose persistence exceeds the relative threshold $\tau$ 
($P \geq \tau$) are categorized as \emph{Persistent recovery} (C3), corresponding 
to sustained and temporally consolidated structural changes.

The threshold $\tau$ is defined as a fixed proportion (60\%) of the available 
temporal frames, ensuring adaptability to different satellite acquisition schedules and recovery observation windows. This relative formulation allows consistent application 
of the framework across cities characterized by heterogeneous temporal depth.

It is important to emphasize that these recovery classes do not directly measure 
construction status or damage severity. Rather, they represent statistically 
consistent departures from the learned post-event SAR dynamics. In this sense, 
the resulting map should be interpreted as a proxy of recovery-related activity 
intensity inferred from temporal persistence patterns.

\subsection{Experimental Setup and Implementation Details}
\label{sec:experimental_setup}
All experiments follow the LOCO protocol described earlier. For each city-fold, the multi-temporal CSK stacks are partitioned into overlapping patches of size $128 \times 128$ pixels using a sliding-window scheme with stride $s=64$, ensuring spatial coverage and contextual redundancy. 
The ConvLSTM autoencoder parameters $\theta$ are trained by minimizing the reconstruction loss defined in Eq.~\eqref{eq:total_loss_nomask} using Adam with learning rate $\eta = 10^{-4}$ and mini-batch size $B=8$. 

\begin{table}[t]
\centering
\caption{Operational definition of SAR-based recovery classes.}
\label{tab:recovery_classes}

\resizebox{\columnwidth}{!}{%
\begin{tabular}{llll}
\toprule
\textbf{Class} & \textbf{ID} & \textbf{Persistence rule} & \textbf{Interpretation} \\
\midrule
Transient activity & C1 & $P = 1$ 
& Isolated / short-lived anomaly \\

Emerging recovery & C2 & $2 \leq P < \tau$ 
& Intermediate, ongoing transformation \\

Persistent recovery & C3 & $P \geq \tau$ 
& Sustained structural change \\

\bottomrule
\end{tabular}
}

\end{table}

\section{Results}
\label{sec:results}

\subsection{Cross-city heterogeneity of recovery patterns}
\label{subsec:cross_city_heterogeneity}

\begin{figure*}
    \centering
    \includegraphics[width=\textwidth]{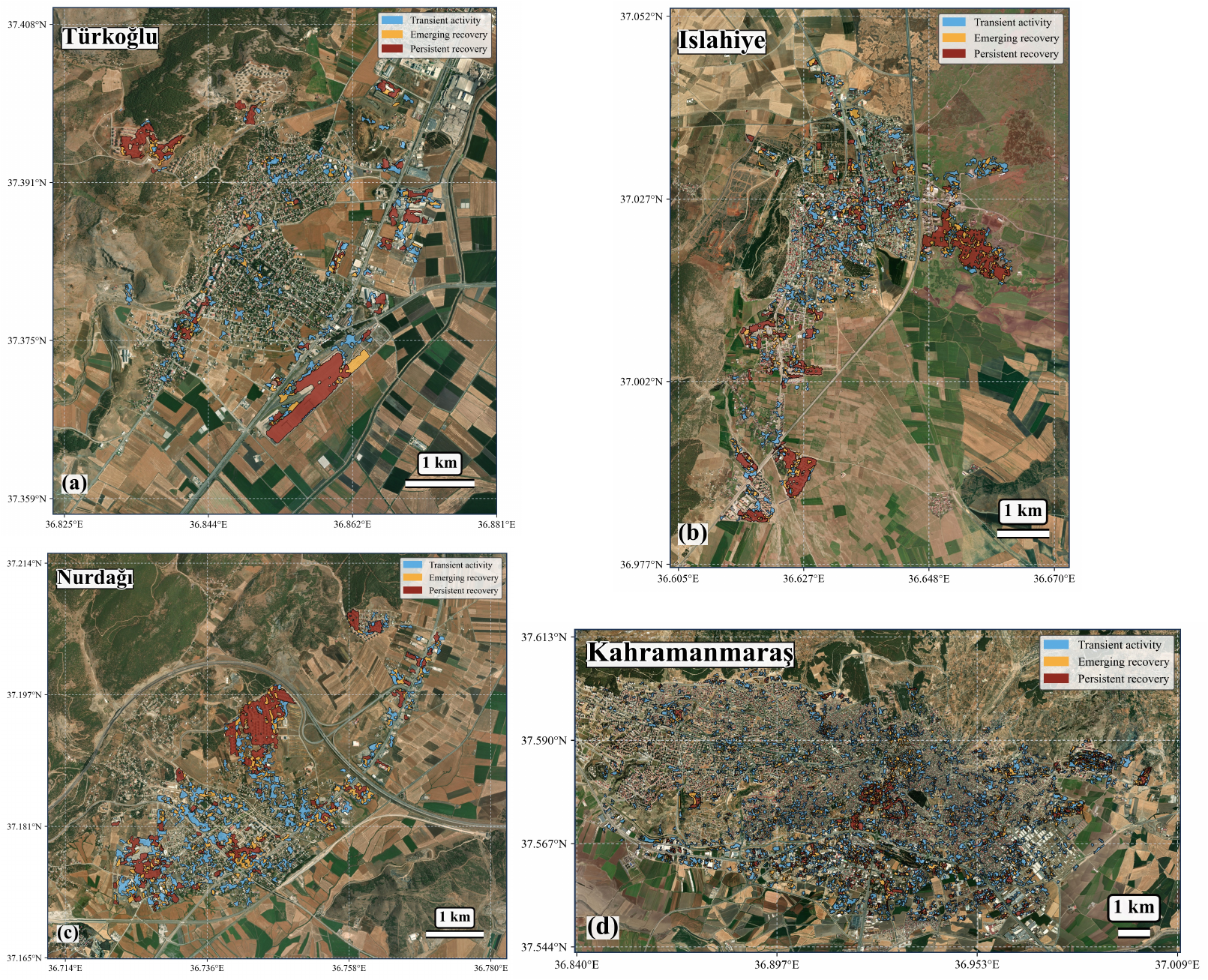}
    \caption{
    Spatial distribution of post-earthquake recovery dynamics derived from multi-temporal SAR anomaly persistence in four affected cities: (a) Türkoğlu, (b) Islahiye, (c) Nurdağı, and (d) Kahramanmaraş. Recovery patterns are categorized into transient activity (light blue), representing short-lived anomaly detections within the time series; emerging recovery (orange), indicating intermediate levels of anomaly persistence; and persistent recovery (red), corresponding to areas exhibiting sustained anomalous behavior across multiple acquisitions. These classes reflect the statistical temporal consistency of SAR-derived anomalies rather than direct observations of reconstruction status. Smaller municipalities exhibit localized clusters of persistent signals, whereas the metropolitan-scale urban fabric of Kahramanmaraş shows more spatially heterogeneous and widespread anomaly patterns.
    }
    \label{fig:results_anomaly_maps}
\end{figure*}

The recovery maps for the four study cities are reported in Fig.~\ref{fig:results_anomaly_maps}. A first qualitative inspection reveals a marked cross-city heterogeneity in both the spatial concentration and the spatial organization of recovery-related signals. At the same time, a common pattern emerges across all cities, and especially across the three smaller urban centers, namely the presence of spatially clustered recovery-related signals rather than isolated and scattered detections. These clustered signals are, in many cases, located outside the historical urban core or along its margins. Although these maps do not directly represent the reconstruction status itself, but rather the temporal consistency of SAR-detected changes, this clustering pattern suggests that organized post-event transformations were already taking place within the first weeks to months after the earthquake.

In the three smaller municipalities (Türkoğlu, Islahiye, and Nurdağı), recovery activity is predominantly organized into a limited number of spatially coherent hotspots. In Türkoğlu, persistent and emerging patterns appear as compact clusters, with prominent concentrations along the urban fringe and near major transportation and industrial corridors, while transient activity remains scattered within the built-up core.

In Islahiye, recovery-related signals are organized along the elongated urban structure of the city, which develops primarily along a north–south corridor. Rather than forming a single dominant hotspot, anomaly patterns are distributed across the built-up fabric, with a well-defined persistent cluster emerging on the eastern side. Additional localized patches of emerging and persistent recovery are visible in peripheral sectors, particularly toward the southern margins, while the central urban area exhibits a more fragmented mixture of transient and intermediate signals. This configuration suggests a spatially distributed and progressively consolidated reconstruction process.

In Nurdağı, recovery-related patterns are concentrated in a limited number of well-defined hotspots, with the most persistent signals forming two major clusters located north of the main built-up area and along its southwestern margin. Emerging recovery surrounds these core zones and extends along nearby urban edges, while transient activity is more diffusely distributed within the central built-up fabric. Overall, this spatial configuration suggests that the most temporally consistent post-event transformations are associated with a small number of concentrated intervention areas, rather than with a uniform reorganization of the entire urban core.

In contrast, Kahramanmaraş exhibits a substantially different recovery configuration. Recovery-related signals are widespread and highly fragmented, forming a heterogeneous mosaic across the metropolitan-scale built-up area. Rather than being dominated by a small number of compact hotspots, the map reveals numerous spatially dispersed activations with varying persistence levels, indicating a spatially heterogeneous reconstruction process unfolding across multiple neighborhoods.

These differences underline how recovery dynamics are not only temporally variable, but also strongly conditioned by urban scale and spatial structure, with distinct configurations emerging across cities of different size and morphology.

\subsection{Class-wise distribution of recovery persistence across cities}
\label{subsec:class_distribution_radar}
\begin{figure}
\centering
\includegraphics[width=0.6\columnwidth]{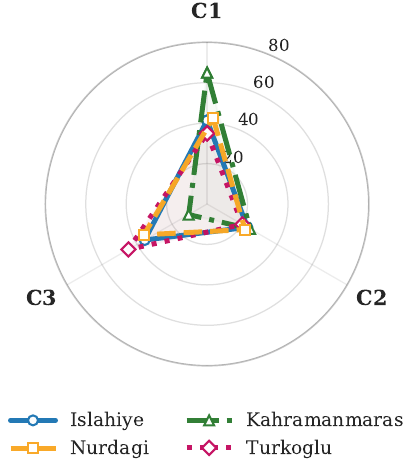}
\caption{
Proportional distribution of persistence-based recovery classes across the four study cities, shown using a radar plot. The axes represent the percentage of the SAR analysis area assigned to transient activity (C1), emerging recovery (C2), and persistent recovery (C3), respectively.
}
\label{fig:recovery_radar}
\end{figure}

Fig.~\ref{fig:recovery_radar} reports the percentage distribution of the three persistence-based recovery classes for each city.

The radar plot highlights a clear structural similarity among the three smaller municipalities 
(Türkoğlu, Islahiye, and Nurdağı). 
In these cities, the relative proportions of transient activity (C1) and persistent recovery (C3) are comparable, 
while emerging recovery (C2) consistently represents a smaller fraction of the total anomalous area. 
This balanced distribution is consistent with the hotspot-driven configurations observed in the maps, 
where compact clusters of sustained transformations coexist with localized short-lived activations.

In contrast, Kahramanmaraş exhibits a markedly different class distribution. 
The transient activity class (C1) dominates the radar profile, accounting for the majority of anomalous pixels, 
whereas persistent recovery (C3) represents only a limited fraction of the total. 
This quantitative imbalance mirrors the spatially fragmented and metropolitan-scale recovery configuration described previously. 
Rather than being structured around a limited number of coherent hotspots, the recovery process in Kahramanmaraş appears 
distributed across numerous neighborhoods, producing widespread but less temporally consolidated anomaly activations.

The radar plot therefore confirms, in aggregate statistical form, the cross-city heterogeneity already evident in the spatial maps: 
while smaller municipalities display a persistence structure characterized by balanced transient and persistent components, 
the metropolitan case is dominated by diffuse, short-lived activations, indicative of a more spatially fragmented recovery process.

\subsection{Temporal expansion of cumulative anomalous area}
\label{subsec:cumulative_recovery_dynamics}

\begin{figure}
\centering
\includegraphics[width=0.9\columnwidth]{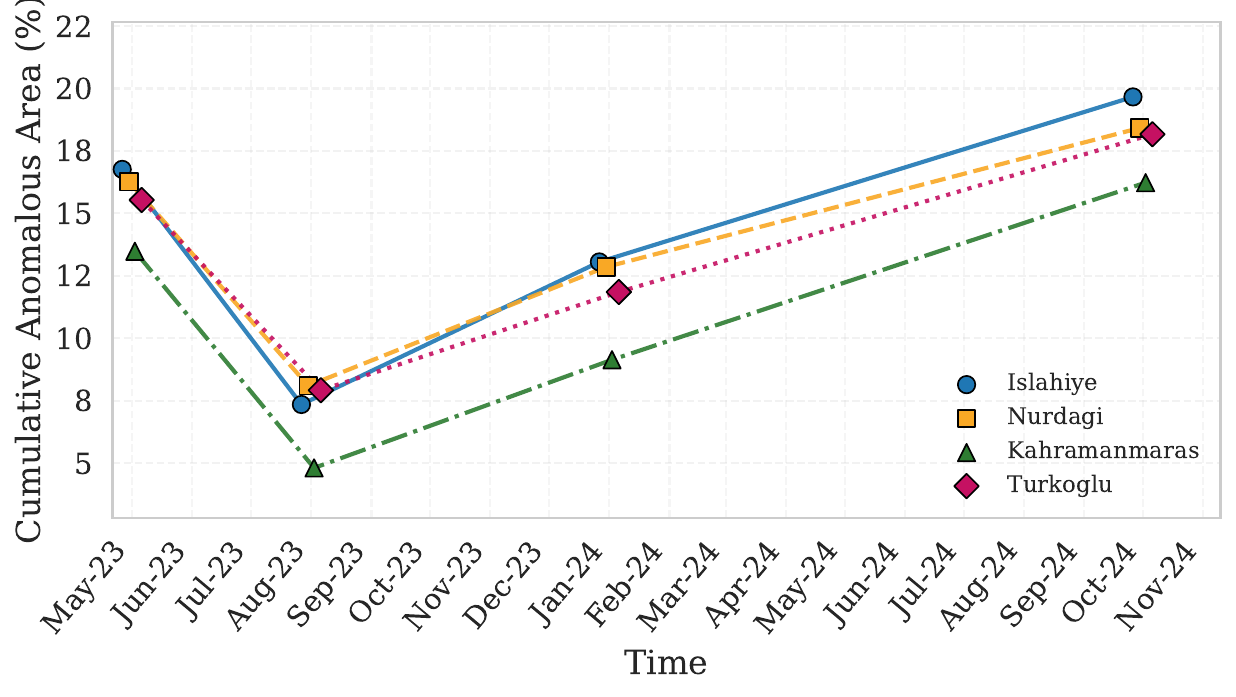}
\caption{
Temporal evolution of cumulative anomalous area (\%) derived from multi-temporal SAR anomaly persistence across the four study cities. 
For each sensing period, the curve represents the percentage of pixels that have been classified as anomalous at least once up to that date, normalized with respect to the total spatial extent of the SAR analysis area. 
The plot highlights cross-city differences in the temporal evolution and magnitude of SAR-derived recovery-related change.
}
\label{fig:cumulative_recovery_4_cities}
\end{figure}

Fig.~\ref{fig:cumulative_recovery_4_cities} provides a complementary temporal perspective by quantifying the cumulative anomalous area over selected sensing periods.

To capture the main phases of the post-earthquake recovery trajectory while maintaining a compact temporal representation, the cumulative metric is evaluated at four representative sensing periods: May 2023 (3 months since the earthquake), August 2023 (6 months), January 2024 (11 months), and October 2024 (20 months). These dates correspond to key stages of the recovery process, spanning the early post-event stabilization phase, the intermediate reconstruction phase, and the longer-term consolidation of rebuilding activities. For each target date, the cumulative anomalous area is computed by aggregating all anomaly detections observed from the beginning of the monitoring period up to the corresponding acquisition.

Importantly, this metric does not distinguish between persistence levels (C1–C3), but rather captures the overall territorial footprint of SAR-detected change within the predefined SAR processing extent.

A common pattern is observed across all cities: a temporary stabilization between May and August 2023 is followed by a steady and pronounced increase through January and October 2024. The early stabilization reflects the attenuation of short-lived anomaly activations after the immediate post-event phase, while the subsequent growth indicates the progressive spatial expansion of recovery-related transformations.

Despite this shared temporal evolution, inter-city differences remain evident. Islahiye consistently exhibits the highest cumulative anomalous area, reaching nearly 20\% of the analyzed urban footprint by October 2024. Nurdağı and Türkoğlu follow similar trajectories, with slightly lower but comparable cumulative values. In contrast, Kahramanmaraş displays systematically lower cumulative percentages across all sensing periods, namely 13.48\% in May 2023, 4.80\% in August 2023, 9.13\% in January 2024, and 16.23\% in October 2024.

This analysis further suggests that recovery dynamics in Kahramanmaraş are spatially more fragmented and less concentrated in large, repeatedly activated clusters. In the smaller municipalities, instead, cumulative growth is more directly associated with the consolidation of well-defined hotspots, leading to a larger proportion of the urban processing extent being progressively affected by persistent anomaly signals.

\subsection{Local-scale validation examples across cities}
\label{subsec:local_validation_examples}

\begin{figure*}
\centering
\includegraphics[width=\textwidth]{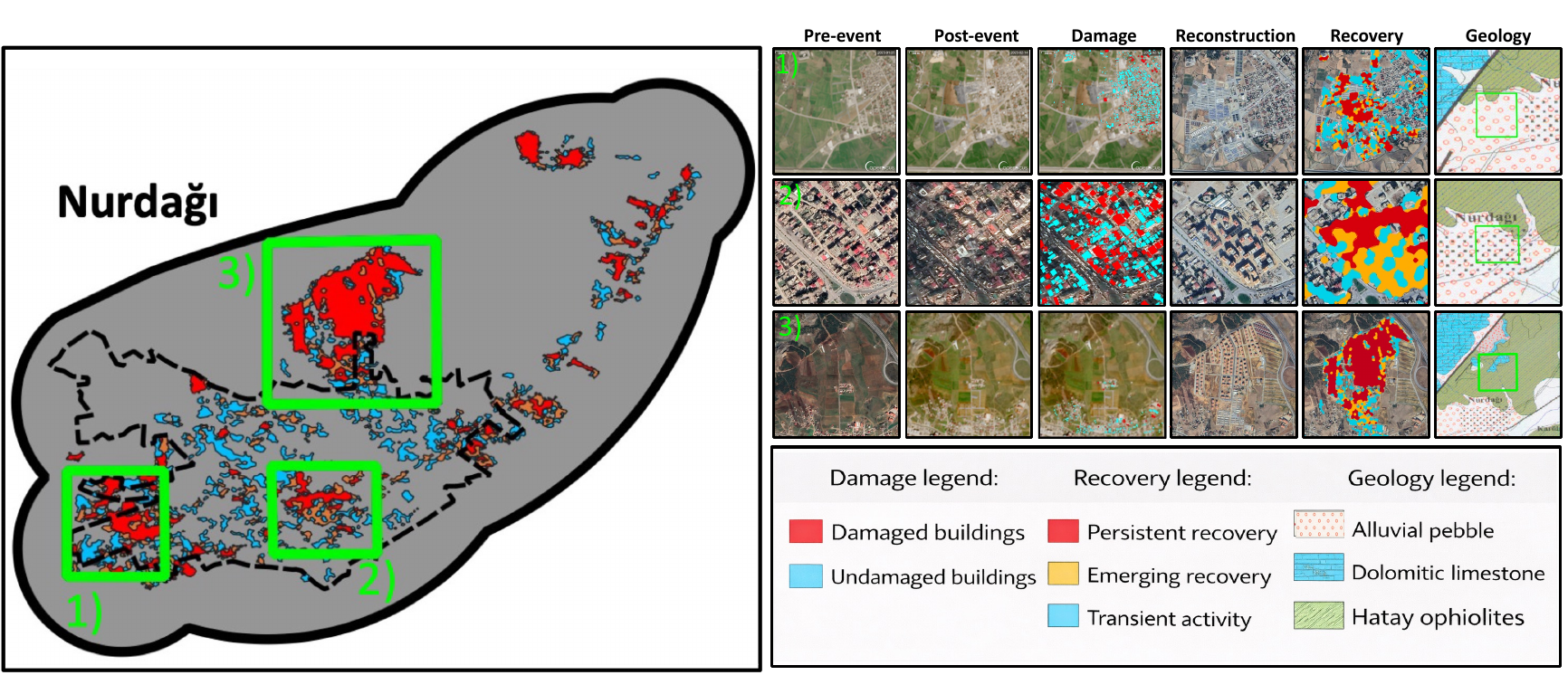}
\caption{
Example of local recovery configurations in three sectors in Nurdağı. Each row shows the temporal evolution of a selected area, including pre-event, post-event (with damage map overlay), reconstruction phase (with recovery map overlay), and the corresponding geological conditions. 
Row 1: pre-event (25-01-2023), post-event and damage map (14-02-2023), reconstruction and recovery map (12-10-2024). 
Row 2: pre-event (20-03-2022), post-event and damage map (07-02-2023), reconstruction and recovery map (12-10-2024). 
Row 3: pre-event (20-03-2022), post-event and damage map (24-02-2023), reconstruction and recovery map (12-10-2024).
Background imagery includes images accessed from Google Earth Pro and Copernicus Sentinel-2.
}
\label{fig:nurdagi}
\end{figure*}

\begin{figure*}
\centering
\includegraphics[width=\textwidth]{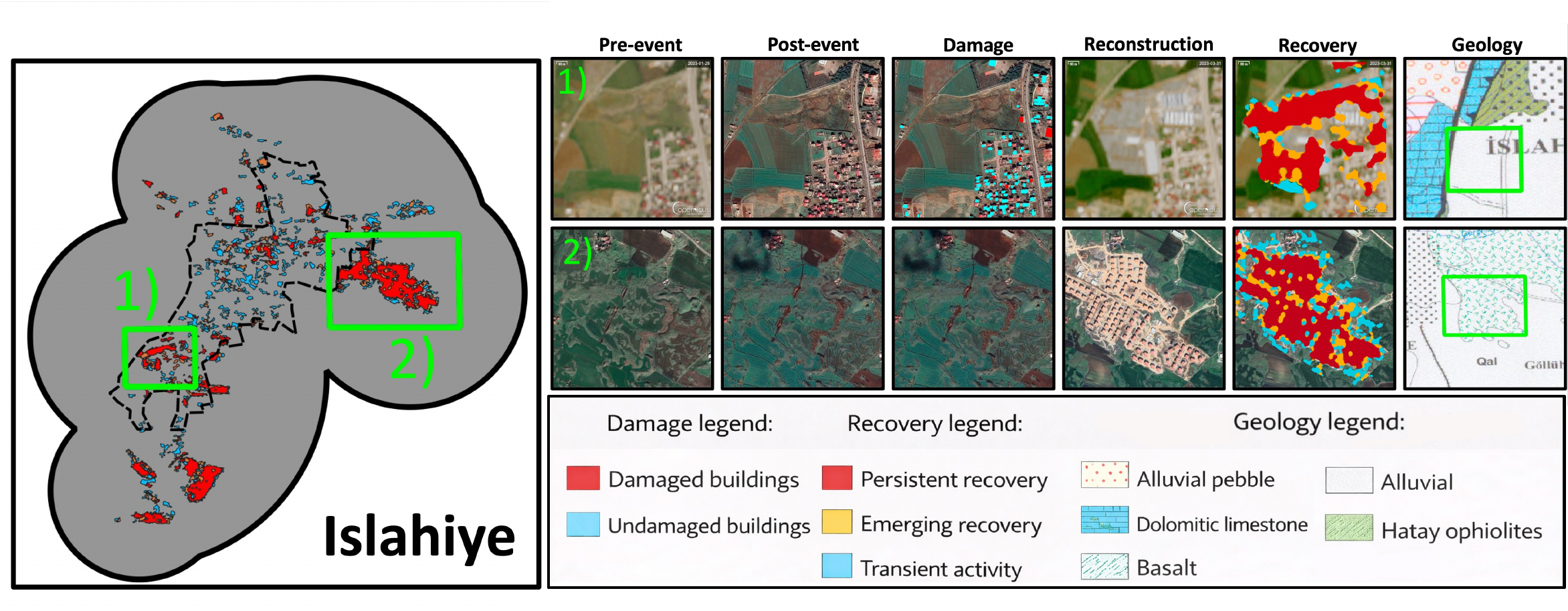}
\caption{
Example of local recovery configurations in Islahiye. Each row shows the temporal evolution of a selected area, including pre-event, post-event (with damage map overlay), reconstruction phase (with recovery map overlay), and the corresponding geological conditions. 
Row 1: pre-event (25-01-2023), post-event and damage map (07-02-2023), intermediate reconstruction image (31-03-2023), and reconstruction and recovery map (05-04-2024). 
Row 2: pre-event (27-12-2022), post-event and damage map (07-02-2023), reconstruction and recovery map (05-04-2024).
Background imagery includes images accessed from Google Earth Pro and Copernicus Sentinel-2.
}
\label{fig:islahiye}
\end{figure*}

\begin{figure*}
\centering
\includegraphics[width=\textwidth]{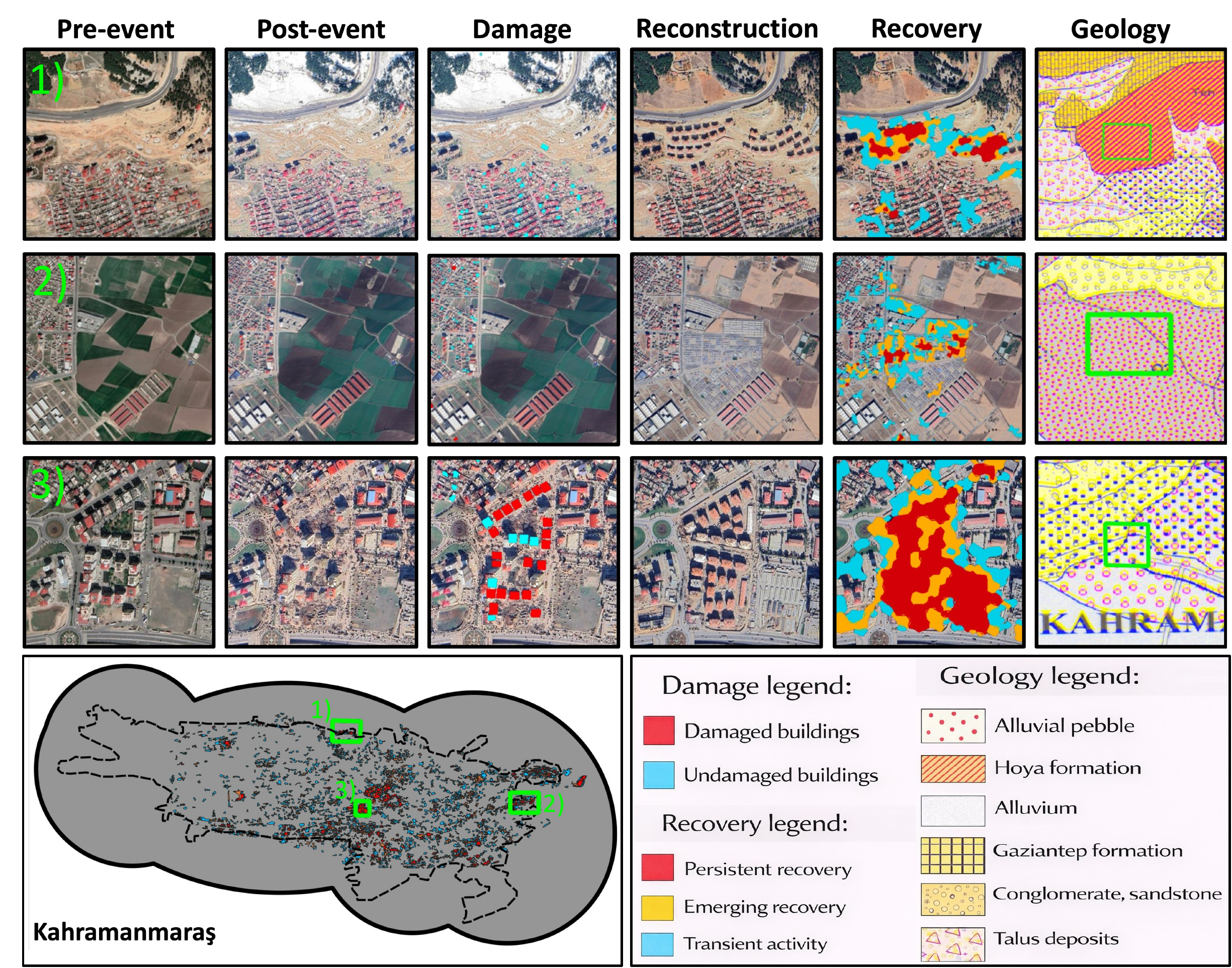}
\caption{
Example of local recovery configurations in Kahramanmaraş. Each row shows the temporal evolution of a selected area, including pre-event, post-event (with damage map overlay), reconstruction phase (with recovery map overlay), and the corresponding geological conditions. 
All rows share the same acquisition dates: pre-event (10-04-2022), post-event and damage map (08-02-2023), and reconstruction and recovery map (30-10-2024).
Background imagery includes images accessed from Google Earth Pro and Copernicus Sentinel-2.
}
\label{fig:kahramanmaras}
\end{figure*}

\begin{figure*}
\centering
\includegraphics[width=\textwidth]{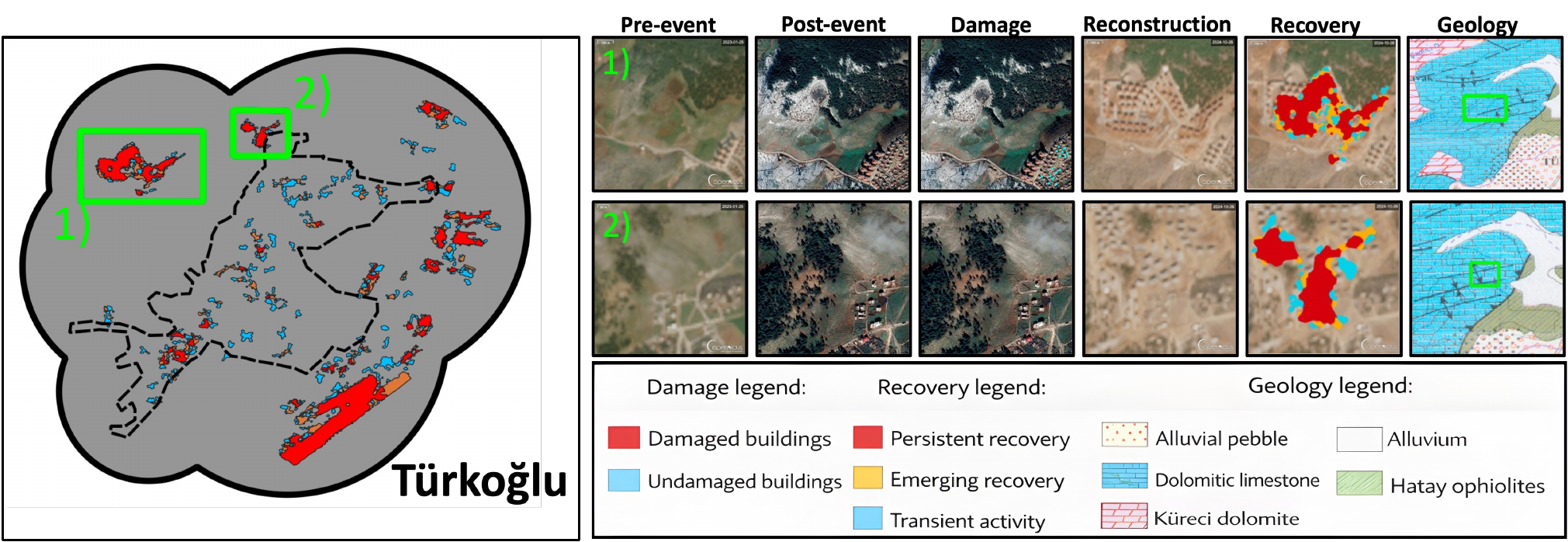}
\caption{
Example of local recovery configurations in Türkoğlu. Each row shows the temporal evolution of a selected area, including pre-event, post-event (with damage map overlay), reconstruction phase (with recovery map overlay), and the corresponding geological conditions. 
Both rows share the same acquisition dates: pre-event (25-01-2023), post-event and damage map (07-02-2023), reconstruction and recovery map (26-10-2024).
Background imagery includes images accessed from Google Earth Pro and Copernicus Sentinel-2.
}
\label{fig:turkoglu}
\end{figure*}

To validate the SAR-derived recovery maps produced by the proposed framework, 
Figs.~\ref{fig:nurdagi}–\ref{fig:turkoglu} display representative local-scale examples for the four study cities, arranged in a consistent six-column layout. The cities are shown in the following order: Nurdağı, Islahiye, Kahramanmaraş, and Türkoğlu.

For each example, the columns follow a fixed sequence: (i) pre-event optical imagery, (ii) post-event optical imagery, (iii) building-damage map, (iv) reconstruction-phase optical imagery, (v) SAR-derived recovery map, and (vi) geological context. This standardized layout ensures a direct and consistent visual comparison across all case studies.

The pre- and post-event images provide a baseline for identifying damage patterns, while the building-damage map highlights areas affected by structural collapse. The reconstruction-phase imagery captures the subsequent urban transformation, including debris removal and the emergence of new structures. The SAR-derived recovery map then identifies persistent anomaly patterns associated with reconstruction activity, enabling a temporally integrated interpretation of the recovery process. Finally, the geological map provides information on local lithological conditions, allowing reconstruction patterns to be interpreted in relation to ground stability and seismic susceptibility.

High-resolution Google Earth imagery was used, whenever available, to support the visual interpretation of pre-event, post-event, and reconstruction phases. In periods characterized by sparse or outdated Google Earth coverage, Copernicus Sentinel-2 time series were used to track rapid urban transformations, such as debris removal, temporary container settlements, and prefabricated housing installations, thereby bridging temporal gaps in the optical validation.

MTA geological maps were incorporated to distinguish reconstruction occurring on mechanically competent lithotypes (e.g., basalt, ophiolites, limestone) from rebuilding on softer alluvial deposits. This information supports the interpretation of recovery patterns in relation to local ground conditions and geotechnical suitability.

Taken together, these examples illustrate the spatial correspondence between SAR-derived anomaly persistence, independently observed reconstruction processes, and the underlying geological setting. In several cases, permanent reconstruction takes place within previously damaged sectors located on alluvial deposits, i.e., within the original urban footprint after collapse, debris removal, and site clearing (see, for example, the second row of Fig.~\ref{fig:nurdagi} and later description). At the same time, newly developed residential districts frequently appear outside the original urban footprint and are often located on mechanically more competent lithologies, such as limestone, basalt, or ophiolitic units.

While it is beyond the scope of the present research assessing whether these reconstruction activities were intentionally designed to more stable geology and thus stating that the reconstruction made the urban building stock more geological resilient, we focus the analysis on demonstrating that multi-temporal SAR-based change maps can serve as an additional geospatial layer for investigating the relationship between reconstruction dynamics and environmental factors, including geological susceptibility.

\paragraph{Nurdağı}\label{sec:nurdagi_description}
Fig.~\ref{fig:nurdagi} reports three representative local-scale examples for Nurdağı.

The first example (row 1) illustrates the rapid deployment of temporary container settlements shortly after the 7 February 2023 earthquake. 
The comparison between pre- and post-event imagery shows that the area was largely undeveloped before the event, while the reconstruction-phase image reveals the rapid installation of regularly spaced prefabricated units. 
The recovery map highlights the central portion of this sector as a persistent anomaly (C3), indicating sustained SAR backscatter changes associated with the presence of these structures. 
The persistence of this signal, still visible in 2024 imagery, suggests that the temporary settlement remains active over an extended period. 
The geological map indicates that the area is located on alluvial deposits, although in this case the ground conditions are less relevant due to the temporary nature of the structures.

The second example (row 2) shows in-situ reconstruction within a previously damaged urban sector. 
The damage map confirms widespread building collapse immediately after the event, which is clearly visible when comparing pre- and post-event imagery. 
The reconstruction-phase image reveals rebuilding activity occurring within the same footprint, while the recovery map identifies this area as persistently anomalous, reflecting ongoing structural transformation. 
The geological context indicates that this sector is predominantly characterized by alluvial deposits, suggesting that reconstruction is taking place in a geotechnically vulnerable setting where damage was initially concentrated. In this case, the building code used for such new permanent constructions will certainly play a role to ensure future seismic resilience.

The third example (row 3) depicts a newly developed residential area located outside the historical urban core. 
The pre-event imagery shows an undeveloped or sparsely used area, while the post-event and reconstruction-phase images document the progressive construction of organized housing blocks. 
The recovery map highlights this sector as persistently anomalous, consistent with large-scale and sustained construction activity. 
In contrast to the previous case, the geological map indicates that this area is located on more mechanically competent lithologies, including dolomitic limestone and ophiolitic units, which provide stiffer and more stable ground conditions compared to the alluvial sediments underlying much of the original city.

\paragraph{Islahiye}
The first example in Fig.~\ref{fig:islahiye} (first row) shows the rapid establishment of temporary container settlements immediately after the event. 
The pre- and post-event images highlight the abrupt appearance of new structures, while the damage map confirms significant disruption in the area. 
The reconstruction image shows the progressive organization of the settlement, and the recovery map identifies this sector as persistently anomalous, indicating sustained structural changes over time. 
The geological layer suggests that these developments occur over alluvial deposits, consistent with the pre-existing urban setting.
The second example (second row) illustrates organized post-event residential development in a previously undeveloped area outside the original urban core. 
The pre-event image shows largely unbuilt land, while the post-event image and the damage map indicate that the area was still mostly undeveloped and not affected by significant building damage. 
The reconstruction image reveals the emergence of a structured residential layout, which is clearly highlighted as persistently anomalous in the recovery map, reflecting long-lasting construction activity. 
The geological map indicates that this sector is located on basaltic lithology, contrasting with the predominantly alluvial deposits of the original urban area, and suggesting a shift of development toward geologically distinct zones.

\paragraph{Kahramanmaraş}
The first example in Fig.~\ref{fig:kahramanmaras} (first row) illustrates the development of a residential area located outside the main pre-event urban fabric. 
The comparison between pre- and post-event imagery shows that this sector does not exhibit significant visible changes immediately after the earthquake, and the building damage map from~\cite{Russo2026BDA} does not indicate substantial structural damage within this area. 
The reconstruction-phase image reveals the subsequent emergence of organized housing blocks, indicating that the development occurred after the initial post-event phase rather than as a direct consequence of damage. 
The recovery map correctly identifies this sector as persistently anomalous, capturing the sustained construction activity over time. 
The geological layer indicates that the area is located on relatively competent lithologies, which may have contributed to the limited observed damage compared to softer deposits in the urban core.

The second example (second row) documents the installation of a large-scale prefabricated settlement on previously agricultural land. 
The comparison between pre- and post-event imagery shows limited immediate change in the central portion of the area, and the building damage map indicates only localized damage outside the main reconstruction zone. 
The reconstruction-phase image clearly reveals the later development of a regular, grid-based layout of prefabricated housing units and internal roads. 
The recovery map captures this transformation through spatially coherent and persistent anomalies, reflecting the substantial surface modification associated with site preparation and settlement installation. 
The geological layer indicates that this area is located on alluvial deposits, characterized by unconsolidated sediments such as pebbles, sand, silt, and clay. 
These softer materials are generally more susceptible to deformation and seismic amplification, suggesting that site selection in this case was primarily driven by land availability and rapid deployment needs rather than optimal geotechnical conditions.

The third example (third row) illustrates reconstruction occurring within a previously damaged urban sector. 
The comparison between pre- and post-event imagery, together with the building damage map from~\cite{Russo2026BDA}, confirms that this area experienced significant structural damage during the earthquake. 
The reconstruction-phase image shows rebuilding activity taking place within the original urban footprint, while the recovery map highlights this sector as persistently anomalous, indicating ongoing structural transformation. 
The geological layer indicates that this area is located on a heterogeneous substrate, consisting of a mixture of alluvial deposits and more consolidated sedimentary formations. 
This combination of soft and more competent units may have contributed to spatial variability in damage patterns and influences the geotechnical conditions under which reconstruction is taking place.

\paragraph{Türkoğlu}

The two examples shown in Fig.~\ref{fig:turkoglu} illustrate new residential development located outside the pre-event urban core. 
In both cases, the pre- and post-event images show no significant visible changes, and the corresponding building damage maps from~\cite{Russo2026BDA} do not indicate substantial structural damage within these areas. 
The reconstruction images instead reveal the subsequent development of organized residential settlements, indicating that these interventions occurred in locations not directly affected by the earthquake. 
The recovery maps classify these areas as persistently anomalous, capturing sustained construction activity over time. 
The geological context indicates that these sectors are located on relatively stable lithologies, namely dolomitic limestone, which may have favored the selection of these areas for reconstruction.

Notably, persistent anomaly patterns are not concentrated within the historical city center; instead, they predominantly emerge in peripheral sectors, as highlighted by the presented examples. 
This suggests that post-earthquake transformation in Türkoğlu is primarily driven by outward urban expansion rather than in-situ reconstruction within the existing urban fabric.

The spatial correspondence between persistent anomaly signals and new construction areas indicates a relocation process toward geologically more stable terrain, distinct from the softer alluvial deposits characterizing parts of the original urban area. 
This pattern is consistent with a structured and spatially concentrated recovery process.

\subsection{Comparison with Previous Work}
\label{sec:comparison_others}

To further evaluate the results of the proposed SAR-based recovery framework, we compare them with the recent NTL-based recovery analysis presented in \cite{gong2025urban}. 
The comparison focuses on the city of Nurdağı, which represents the only study area shared between the two investigations.

Their approach exploits time-series observations from the SDGSAT-1 satellite to monitor post-disaster recovery through variations in night-time radiance. Recovery dynamics are quantified using the \textit{Power Recovery Percentage} (PRP) metric, defined as the ratio between post-event and pre-event nighttime radiance levels. This indicator measures the relative restoration of night-time illumination compared to pre-disaster conditions and is interpreted as a proxy for the recovery of electricity supply and associated socioeconomic activity. Based on the temporal evolution of PRP values, Gong et al. further identify representative recovery patterns by clustering pixel-level NTL time series.

The analysis presented in \cite{gong2025urban} covers the first year after the earthquake, with the most recent acquisition used for their recovery assessment dated April 3, 2024. 
To ensure temporal consistency between the two analyses,  the SAR-based recovery map used for the comparison with Gong et al. is the one generated from the CSK image collected on January 11, 2024, i.e. the one among the subset dataset used in this research that is the closest in time prior to April 3, 2024.

\begin{figure*}
\centering
\includegraphics[width=\textwidth]{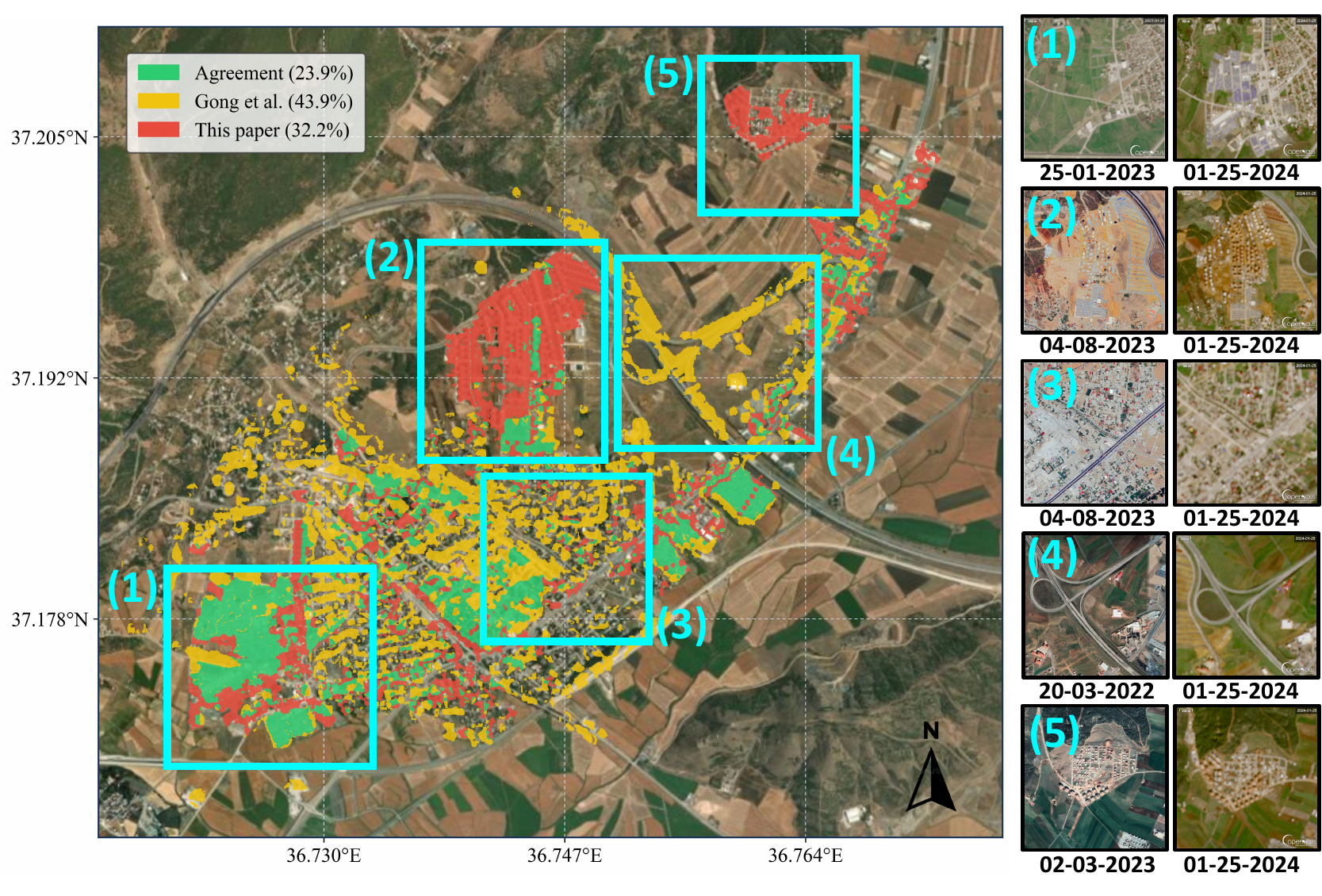}
\caption{
Spatial comparison between recovery signals detected by the proposed SAR-based framework and the NTL-based recovery map from \cite{gong2025urban} for the city of Nurdağı. Pixels are classified as agreement (green), NTL-only detections (yellow), and SAR-only detections (red). Percentages indicate the relative spatial extent of each class within the shared analysis domain. Five representative areas of interest (1–5) highlight different recovery scenarios. The corresponding inset pairs on the right show the temporal evolution of each area between 2022–2023 (pre-/early post-event) and 2024, illustrating reconstruction processes such as temporary settlements, new residential developments, and infrastructure-related signals. Background imagery includes images accessed from Google Earth Pro and Copernicus Sentinel-2.
}
\label{fig:sar_ntl_comparison}
\end{figure*}

Fig.~\ref{fig:sar_ntl_comparison} illustrates the spatial agreement between the two approaches for the city of Nurdağı. 
Three categories are identified: areas where both methods detect recovery (agreement), areas detected only by the NTL-based approach of \cite{gong2025urban}, and areas identified exclusively by the proposed SAR-based framework. 
Within the shared analysis domain, 23.9\% of the pixels show agreement between the two approaches, while 43.9\% are detected only by the NTL-based method and 32.2\% are identified exclusively by the SAR-based framework. 
Although a substantial portion of the recovery patterns is consistently detected by both methods, noticeable differences emerge across several sectors of the urban domain.

A closer inspection of Fig.~\ref{fig:sar_ntl_comparison}, supported by the underlying optical imagery, reveals that part of the NTL-only detections (yellow) corresponds to linear features associated with transportation infrastructure, such as major roads and highway interchanges. These areas are illuminated by street lighting rather than reflecting reconstruction activities within the built environment, and may therefore be incorrectly interpreted as recovery signals in the NTL-based product.

Example (4) clearly illustrates this effect. The inset shows a highway interchange that is already present in the pre-event imagery and remains essentially unchanged in the 2024 observations. Despite the absence of structural transformation, this area is detected as recovery by the NTL-based approach, likely due to persistent or increased night-time illumination. This confirms that NTL observations can produce false positives when changes in lighting conditions are not directly related to reconstruction processes.

At the same time, not all NTL-only detections should be interpreted as errors or unrelated signals. Example (3) represents a case where the NTL-based approach detects recovery while the SAR-based framework does not identify significant anomalies. Optical imagery suggests that this area may have undergone functional reactivation without substantial structural modification, or that changes occurred at a scale or in a form that is less detectable in SAR backscatter. In such cases, NTL observations may capture the early restoration of human activity or electricity supply, even in the absence of major construction.

Conversely, examples (2) and (5) illustrate areas where the proposed SAR-based framework detects clear reconstruction activity that is not captured by the NTL-based recovery map. Example (2) is also reported in Fig.~\ref{fig:nurdagi}. Optical imagery confirms that by early 2024 these locations host newly constructed residential developments and large-scale housing complexes resulting from the reconstruction process. However, the corresponding NTL-based map from \cite{gong2025urban} does not indicate recovery signals in these areas. A plausible explanation is that these newly built neighborhoods were still partially unoccupied or not fully connected to the electricity network at the time of the NTL observation, resulting in limited night-time illumination.

Example (1) highlights a region where both approaches consistently detect recovery signals. This area corresponds to a post-disaster accommodation zone characterized by the rapid deployment of temporary container settlements. As these facilities are physically present and associated with night-time illumination (e.g., due to temporary power supply or emergency lighting), they generate consistent signals in both SAR and NTL observations.

These examples demonstrate that discrepancies between SAR- and NTL-based observations are not solely indicative of errors, but rather reflect different sensitivities to structural and functional dimensions of post-disaster recovery. 
NTL-based methods provide valuable insights into the restoration of socioeconomic activity and electricity supply, while SAR-based approaches are capable of identifying physical reconstruction processes at earlier stages. 
In particular, SAR observations can detect large-scale construction activities and the emergence of new built-up areas even before these locations become functionally active or illuminated at night.

Overall, the comparison indicates that SAR-based monitoring provides an earlier and structurally grounded perspective on post-disaster reconstruction, whereas NTL-based indicators capture the subsequent reactivation of urban functionality. 
The two approaches therefore offer complementary and mutually informative views of recovery dynamics.

\section{Discussions}\label{sec:discussions}

\subsection{Potential of unsupervised SAR-based recovery monitoring}

The results presented in this study demonstrate the potential of multi-temporal SAR observations combined with unsupervised learning to monitor post-disaster recovery processes at the urban scale. In post-disaster contexts, the recovery phase typically unfolds while reconstruction activities are still evolving, and reliable ground-truth information on newly built structures is rarely available. This creates a practical limitation for supervised approaches, which require annotated datasets that are difficult to obtain during the early and intermediate stages of reconstruction. In this perspective, the proposed unsupervised framework reflects a realistic operational scenario, where EO time series can be exploited to detect and track recovery dynamics without relying on labeled training data.

Within this context, SAR observations provide a particularly valuable capability. By exploiting the temporal consistency of anomaly signals derived from CSK time series, the proposed framework enables the identification of spatially heterogeneous reconstruction dynamics and offers a physically grounded perspective on the evolution of urban landscapes following a major seismic event. An additional advantage of SAR data lies in their independence from daylight and weather conditions, allowing systematic monitoring of disaster-affected areas even when optical imagery may be unavailable.

Across the four investigated cities, the resulting recovery maps reveal that reconstruction processes do not occur uniformly across the urban fabric but instead follow spatially structured patterns. Persistent anomaly clusters frequently correspond to areas undergoing large-scale transformations, including temporary container settlements, debris clearance operations, and the construction of new residential districts. Conversely, areas classified as transient activity often reflect localized or short-lived changes, while emerging recovery signals represent intermediate stages of reconstruction where structural transformations progressively stabilize. These observations confirm that recovery processes following large earthquakes are inherently heterogeneous and unfold at different temporal scales across the urban domain.

An additional operational advantage of the proposed approach is related to the availability of SAR observations acquired through dedicated disaster acquisition strategies. In particular, the CSK mission operated by ASI systematically acquires imagery over urbanized areas, high risk and disaster-prone regions and activates priority acquisitions in response to major events. This operational framework ensures rapid data availability and dense temporal coverage, enabling the exploitation of consistent multi-temporal SAR archives not only for immediate damage assessment but also for the longer-term monitoring of reconstruction dynamics.

The comparison with NTL-based recovery products, such as those presented in \cite{gong2025urban}, further highlights the complementary nature of the two observation modalities. NTL indicators primarily capture the restoration of electricity supply and the reactivation of nighttime human activity, which generally occurs once infrastructure services and residential occupancy have been re-established. In contrast, SAR backscatter variations are directly sensitive to physical transformations of the built environment, such as debris removal, surface clearing, and the construction of new buildings. As a result, SAR-derived anomaly signals may anticipate NTL-based recovery indicators, providing an earlier and structurally grounded view of reconstruction processes.

While multi-sensor approaches integrating SAR, optical imagery, and NTL may further refine the interpretation of recovery dynamics, the proposed SAR-based framework already provides a self-contained and operationally robust capability for tracking reconstruction processes using radar observations alone. In this perspective, SAR time series can represent a reliable backbone for post-disaster monitoring, while additional data sources may complement and enrich the interpretation of the detected recovery patterns.

It is important to emphasize that the objective of this analysis is not to prescribe reconstruction strategies, but rather to document the spatial evolution of recovery processes using EO data. Decisions concerning relocation, reconstruction planning, and land-use management are ultimately taken by governmental and planning authorities. However, EO-derived products such as the recovery maps proposed in this study can provide an objective and spatially explicit evidence base that may support these processes.
When combined with geological information, the monitoring of reconstruction dynamics also offers the possibility to evaluate whether permanent rebuilding activities tend to occur on more stable geological units or within areas characterized by higher seismic susceptibility. Temporary emergency settlements are often installed in open and accessible areas that may coincide with alluvial plains or other geologically vulnerable settings. In contrast, long-term reconstruction ideally aims to reduce exposure to unfavorable ground conditions. The ability to document these spatial patterns during the early stages of the recovery phase highlights the potential role of EO-based monitoring in supporting evidence-informed decision making and in promoting the integration of environmental constraints into post-disaster reconstruction planning.

\subsection{Limitations and future research directions}

Despite the promising results obtained in this study, several challenges and limitations should be acknowledged.

First, the proposed framework relies on unsupervised anomaly detection, which identifies deviations from learned temporal patterns rather than explicitly labeling specific reconstruction activities. As a consequence, some anomaly signals may correspond to processes unrelated to post-disaster recovery, such as seasonal land-surface changes, temporary construction activities not associated with earthquake rebuilding, or residual radiometric fluctuations. Although spatial filtering and temporal persistence analysis help mitigate these effects (see for example \cite{velasquez2025monitoring} with regard to amplitude change detection based recovery monitoring), the interpretation of detected anomaly patterns may still benefit from external contextual information, such as optical imagery or ancillary geospatial datasets.

Second, the temporal sampling of SAR acquisitions may limit the capability to fully characterize the detailed trajectory of reconstruction processes. While the CSK time series used in this study captures key phases of post-earthquake recovery, irregular revisit intervals constrain the precise identification of the onset and duration of specific reconstruction activities. Future studies could benefit from denser SAR time series obtained through the integration of observations from multiple radar missions (e.g., Sentinel-1, TerraSAR-X, or future SAR constellations), which would further improve the temporal continuity of recovery monitoring. However, multi-sensor SAR based approaches require that differences in spatial resolution must be handled and addressed, alongside differences in sensitivity and temporal coherence between different radar bands.

Third, the current validation strategy relies primarily on visual interpretation using high-resolution optical imagery. Although this approach provides valuable qualitative confirmation of reconstruction hotspots, the lack of systematic ground-truth datasets describing recovery processes limits the possibility of performing a fully quantitative accuracy assessment. The development of standardized reference datasets for post-disaster recovery monitoring therefore remains an open challenge for the EO community.

Another limitation concerns the intrinsic complexity of SAR backscatter behavior in dense urban environments. Speckle noise, layover effects, and multi-bounce scattering may introduce variability in radar signals that is not directly related to structural changes. While the DL model implicitly learns stable temporal patterns and reduces part of this variability, residual false positives may still occur, particularly in highly heterogeneous urban areas or in zones characterized by dense vegetation.

Finally, the present study focuses primarily on detecting the spatial distribution and temporal persistence of reconstruction-related changes, without explicitly modeling the broader socioeconomic and infrastructural factors that influence recovery dynamics. Post-disaster reconstruction is shaped by multiple interacting drivers, including population displacement, infrastructure restoration, economic resources, and urban planning policies. Integrating EO-derived recovery indicators with socioeconomic datasets and urban morphology metrics could therefore provide a more comprehensive understanding of the mechanisms governing reconstruction processes.

Future research could further extend the proposed framework along several directions. First, while SAR time series already provide a robust and operational backbone for detecting structural transformations of the built environment, the integration of complementary observations such as optical imagery and NTL data could further refine the interpretation of recovery dynamics by jointly capturing both physical reconstruction processes and the progressive reactivation of urban activities.


Moreover, in this study geological information was already incorporated at the interpretation stage, by analysing the recovery-mapping outputs in relation to lithological conditions and the potential geotechnical implications of rebuilding patterns. This analysis suggested that some new residential developments are emerging in areas underlain by more competent lithologies, such as basalt or limestone, while other reconstruction processes remain concentrated within previously damaged sectors on alluvial deposits. Future work could extend this component from qualitative and case-based interpretation to a more explicit and systematic integration of geological and geotechnical information within the recovery-monitoring workflow, for example by quantitatively assessing whether reconstruction preferentially occurs in areas characterized by more stable ground conditions. Finally, the proposed unsupervised framework could be adapted to other disaster contexts, including floods, landslides, and volcanic eruptions, where recovery monitoring remains a largely unexplored application of EO time series, given the very few published papers (see e.g. \cite{de2021monitoring,velasquez2025monitoring}) By enabling automated and scalable detection of long-term landscape transformations, such approaches may contribute to operational disaster-risk management and support broader international resilience and disaster-risk reduction initiatives.

\section{Conclusions}
\label{sec:conclusions}

This study presented an unsupervised framework for monitoring post-disaster recovery dynamics using multi-temporal regular high resolution CSK SAR observations. By exploiting the temporal persistence of anomaly signals extracted from SAR time series through a ConvLSTM-based autoencoder, the proposed approach enables the identification of reconstruction processes without relying on labeled training data, reflecting realistic operational conditions where reliable ground-truth information is often unavailable.

The application of the framework to four cities affected by the 2023 Türkiye–Syria earthquake sequence revealed that recovery processes exhibit strong spatial heterogeneity. In the smaller municipalities, recovery activity tends to concentrate in a limited number of spatially coherent hotspots coinciding with the construction of new residential neighborhoods, whereas the metropolitan-scale urban fabric of Kahramanmaraş shows more fragmented and spatially distributed patterns. Persistent anomaly clusters were frequently associated with reconstruction within previously damaged neighborhoods, the installation of temporary container settlements, and the development of new residential districts. The persistence-based classification further provides an interpretable summary of the temporal behavior of SAR-derived anomalies, distinguishing between transient activity, emerging recovery, and persistent recovery across the urban landscape.

The comparison with NTL-based recovery indicators highlights the complementary nature of SAR and nighttime light observations. While NTL data primarily reflect the restoration of electricity supply and nighttime socioeconomic activity, SAR observations directly capture structural transformations of the built environment, allowing reconstruction processes to be detected at earlier stages and as they progress.

From an application perspective, the recovery maps produced by the proposed framework can be interpreted as a prototype of a thematic EO product for monitoring post-disaster reconstruction dynamics. By providing spatially explicit information on where reconstruction-related changes occur and how they evolve over time, such a product could support public authorities and disaster management agencies in following the progression of recovery processes and performing an ongoing situational assessment of the reconstruction phase. In combination with contextual information such as geological conditions, these EO-derived maps may also facilitate the identification of emerging reconstruction hotspots and the evaluation of their spatial relationship with environmental constraints.

Taken together, these findings highlight the potential of multi-temporal high resolution SAR observations combined with unsupervised learning for monitoring post-disaster reconstruction dynamics and supporting long-term recovery assessment in disaster-affected regions.

\section*{Acknowledgments}
This research was performed in the framework of a PhD funded by NextGenerationEU, Action 4, DM n. 118, 02/03/2023. COSMO-SkyMed Products © of the Italian Space Agency (ASI) were accessed under a license to use granted by ASI within the Project Card TurkeyEQ2024. 

\bibliographystyle{IEEEtran.bst}
\bibliography{references}

\begin{minipage}[t]{0.47\textwidth}
\begin{IEEEbiography}[{\includegraphics[width=1in,height=1.25in,clip,keepaspectratio]{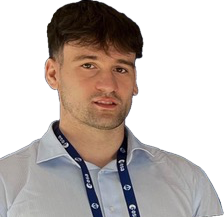}}]{Luigi Russo}
Student Member, IEEE, earned his master's degree (cum laude) in Electronic Engineering for Automation and Telecommunications from the University of Sannio, Benevento, Italy, in 2023. He is currently working toward the Ph.D. degree in Electronics, Computer Science and Electrical Engineering at the University of Pavia, Italy, in collaboration with the Italian Space Agency (ASI), Rome, Italy. He was the winner of the IEEE GRS Italy Chapter Frank Marzano Award for one of the top three Master's theses in Geosciences and Remote Sensing. He has coauthored and presented papers at several prestigious conferences and received the Best Poster Award in Urban and Data Analysis as a young scientist at the 2024 European Space Agency (ESA)-Dragon Symposium. His research interests include the application of artificial intelligence and remote sensing to the analysis of urban and environmental tasks, with particular emphasis on damage mapping and exposure assessment following natural disasters and extreme events.
\end{IEEEbiography}
\end{minipage}

\begin{minipage}[t]{0.47\textwidth}
\begin{IEEEbiography}[{\includegraphics[width=1.05in,height=1.20in,clip,keepaspectratio]{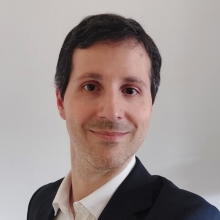}}]{Deodato Tapete} Researcher in Earth Observation and data analytics at the Italian Space Agency (ASI) since 2017, PhD in Earth Sciences, specialised in Synthetic Aperture Radar (SAR) and optical satellite remote sensing for monitoring of cultural heritage, archaeological remote sensing, assessment of natural and anthropogenic hazards, urban applications, agriculture and water resources management. He has developed several methods based on SAR data processing to address issues of heritage conservation, including but not limited to structural stability, impacts of infrastructure construction, weathering and deterioration, looting, intentional destruction. He is ASI program scientist for the Committee on Earth Observation Satellites (CEOS) Working Group on Disaster (WGD). He also leads the ASI programme "Innovation for Downstream Preparation for Science".
\end{IEEEbiography} 
\end{minipage}


\begin{minipage}[t]{0.47\textwidth}
\begin{IEEEbiography}[{\includegraphics[width=1.05in,height=1.20in,clip,keepaspectratio]{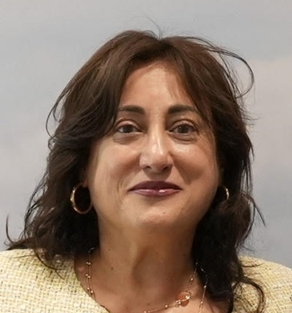}}]{Silvia Liberata Ullo}
IEEE Senior Member, President of IEEE AESS Italy Chapter, Industry Liaison for IEEE Joint ComSoc/VTS Italy Chapter since 2018. Member of the Image Analysis and Data Fusion Technical Committee (IADF TC) of the IEEE Geoscience and Remote Sensing Society (GRSS) since 2020. GRSS Europe Liaison since January 2024. Editor in Chief of IET IMage Processing. Graduated with laude in 1989 in Electronic Engineering at the University of Naples (Italy), pursued the M.Sc. in Management at MIT (Massachusetts Institute of Technology, USA) in 1992. Researcher and teacher since 2004 at University of Sannio, Benevento (Italy). Member of Academic Senate and PhD Professors’ Board. Courses: Signal theory and elaboration, Telecommunication networks (Bachelor program); Earth monitoring and mission analysis Lab (Master program), Optical and radar remote sensing (Ph.D. program). Authored 90+ research papers, co-authored many book chapters and served as editor of two books. Associate Editor of relevant journals (IEEE TGRS, IEEE JSTARS, IEEE GRSL, MDPI Remote Sensing, IET Image Processing, Springer Arabian Journal of Geosciences. Research interests: signal processing, radar systems, sensor networks, smart grids, remote sensing, satellite data analysis, machine learning and quantum ML applied to remote sensing.
\end{IEEEbiography} 
\end{minipage}


\begin{minipage}[t]{0.47\textwidth}
\begin{IEEEbiography}[{\includegraphics[width=1in,height=1.25in,clip,keepaspectratio]{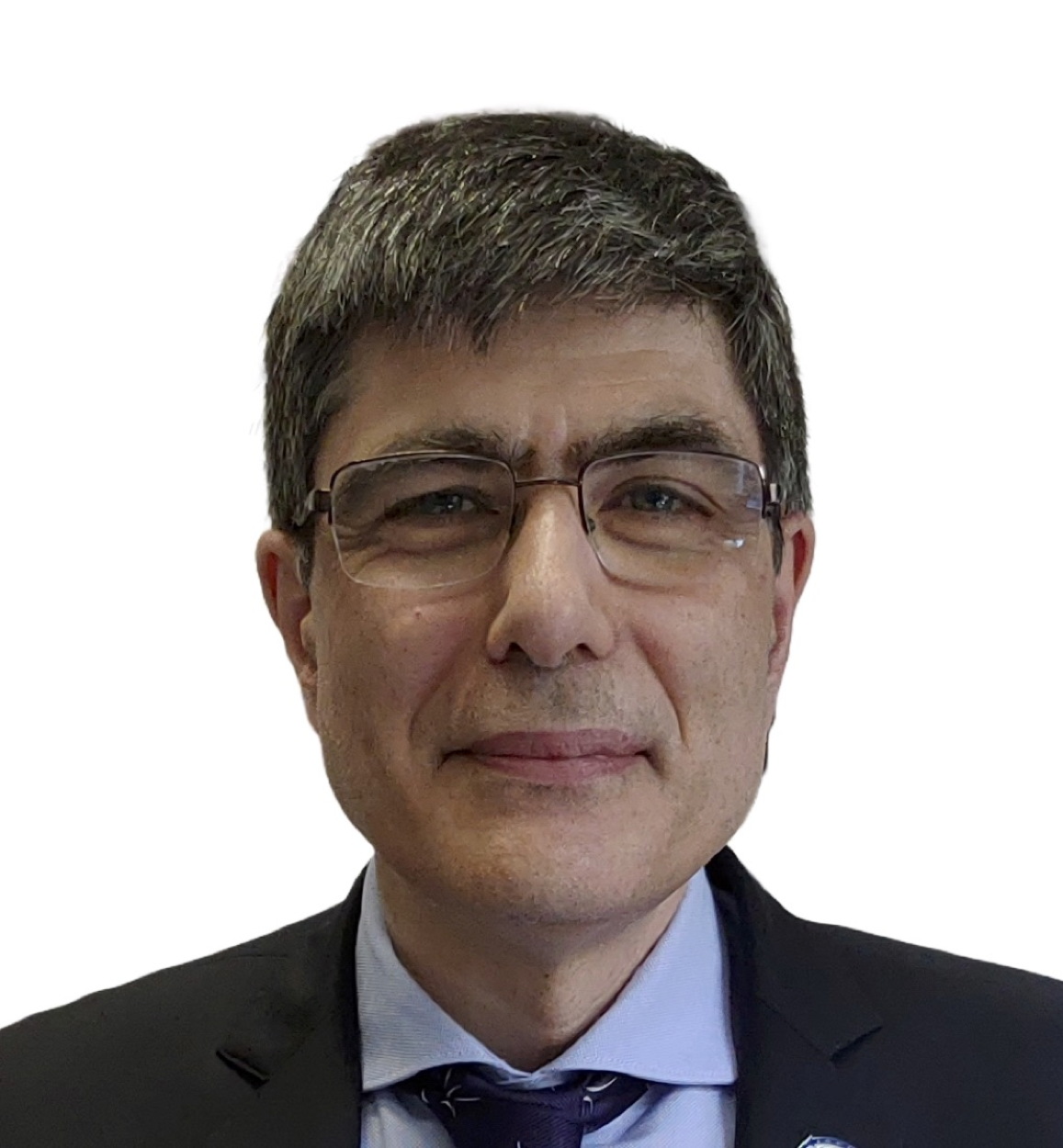}}]{Paolo Gamba} is Professor at the University of Pavia, Italy in the Telecommunications and Remote Sensing Laboratory. He received the Laurea degree in Electronic Engineering “cum laude” from the University of Pavia, Italy, in 1989, and the Ph.D. in Electronic Engineering from the same University in 1993.
He served as Editor-in-Chief of the IEEE Geoscience and Remote Sensing Letters from 2009 to 2013 and of the IEEE Geoscience and Remote Sensing Magazine in 2023-2024. He was Chair of the Data Fusion Committee of the IEEE Geoscience and Remote Sensing Society (GRSS) from October 2005 to May 2009. He has been elected in the GRSS AdCom from 2014 to 2022 and served as GRSS President from 2019 to 2020. He also served as Technical Co-Chair of the 2010, 2015 and 2020 IGARSS conferences, in Honolulu (Hawaii), Milan (Italy), and on-line, respectively.
He is Fellow of IEEE, IAPR, AAIA and the Academia Europaea. He has been invited to give keynote lectures and tutorials on several occasions about urban remote sensing, data fusion, EO data for physical exposure and risk management. He published more than 210 papers in international peer-review journals.
\end{IEEEbiography}
\end{minipage}

\end{document}